\documentclass[conference]{IEEEtran}
\IEEEoverridecommandlockouts
\usepackage{cite}
\usepackage{amsmath,amssymb,amsfonts}
\usepackage{algorithmic}
\usepackage{graphicx}
\usepackage{textcomp}
\usepackage{xcolor}
\usepackage{gensymb}
\usepackage{mathpazo}
\usepackage{algorithm}
\usepackage{esvect}
\usepackage{amsmath}
\usepackage{tikz}
\usepackage{subfig}
\usepackage{hyperref}
\def\BibTeX{{\rm B\kern-.05em{\sc i\kern-.025em b}\kern-.08em
    T\kern-.1667em\lower.7ex\hbox{E}\kern-.125emX}}
\begin{document}

\title{Extended Object Tracking in Curvilinear Road Coordinates for Autonomous Driving 
\thanks{Supported by project TEINVEIN: TEcnologie INnovative per i VEicoli Intelligenti, CUP (Codice Unico Progetto - Unique Project Code): E96D17000110009 - Call ``Accordi per la Ricerca e l'Innovazione", cofunded by POR FESR 2014-2020 (Programma Operativo Regionale, Fondo Europeo di Sviluppo Regionale – Regional Operational Programme, European Regional Development Fund).}
}

\author{
\IEEEauthorblockN{Pragyan Dahal}
\IEEEauthorblockA{\textit{Dept. of Mechanical Engeneering} \\
\textit{Politecnico di Milano}\\
Milano, Italy \\
pragyan.dahal@polimi.it}\and
\IEEEauthorblockN{Simone Mentasti}
\IEEEauthorblockA{\textit{Dept. of Electronics, Information} \\ \textit{ and Bioengineering} \\
\textit{Politecnico di Milano}\\
Milano, Italy \\
simone.mentasti@polimi.it}\and
\IEEEauthorblockN{Stefano Arrigoni}
\IEEEauthorblockA{\textit{Dept. of Mechanical Engeneering} \\
\textit{Politecnico di Milano}\\
Milano, Italy \\
stefano.arrigoni@polimi.it}\and
\IEEEauthorblockN{Francesco Braghin}
\IEEEauthorblockA{\textit{Dept. of Mechanical Engeneering} \\
\textit{Politecnico di Milano}\\
Milano, Italy \\
francesco.braghin@polimi.it}\and
\IEEEauthorblockN{Matteo Matteucci}
\IEEEauthorblockA{\textit{Dept. of Electronics, Information} \\ \textit{ and Bioengineering} \\
\textit{Politecnico di Milano}\\
Milano, Italy \\
matteo.matteucci@polimi.it}\and
\IEEEauthorblockN{Federico Cheli}
\IEEEauthorblockA{\textit{Dept. of Mechanical Engeneering} \\
\textit{Politecnico di Milano}\\
Milano, Italy \\
federico.cheli@polimi.it}\

}

\maketitle
\thispagestyle{plain}
\pagestyle{plain}

\begin{abstract}
In literature, Extended Object Tracking (EOT) algorithms developed for autonomous driving predominantly provide obstacles state estimation in cartesian coordinates in the Vehicle Reference Frame. However, in many scenarios, state representation in road-aligned curvilinear coordinates is preferred when implementing autonomous driving subsystems like cruise control, lane-keeping assist, platooning, etc. This paper proposes a Gaussian Mixture Probability Hypothesis Density~(GM-PHD) filter with an Unscented Kalman Filter~(UKF) estimator that provides obstacle state estimates in curvilinear road coordinates. We employ a hybrid sensor fusion architecture between Lidar and Radar sensors to obtain rich measurement point representations for EOT. The measurement model for the UKF estimator is developed with the integration of coordinate conversion from curvilinear road coordinates to cartesian coordinates by using cubic hermit spline road model. The proposed algorithm is validated through Matlab Driving Scenario Designer simulation and experimental data collected at Monza Eni Circuit. \textit{The Experimental Dataset will be made publicly available upon the paper acceptance}
 \end{abstract}


\section{Introduction}
Correct perception of the ego vehicle's surrounding environment is central in implementing autonomous driving. However, the dynamic and uncertain nature of the environment makes this task very challenging. As the ego vehicle perceives the environment through sensors like Camera, Lidar, and Radar, the data collected from these sensors need to be processed to an understandable format to make driving decisions. The driving software's detection and tracking blocks consistently provide the state of various objects of interest (e.g., vulnerable road users, road traffic participants, etc.) to the decision stack. Our work focus on obstacle tracking with two of the most used sensors in autonomous driving, Lidar, and Radar.

\begin{figure}[t]
    \centering
    \includegraphics[width = 0.48\textwidth]{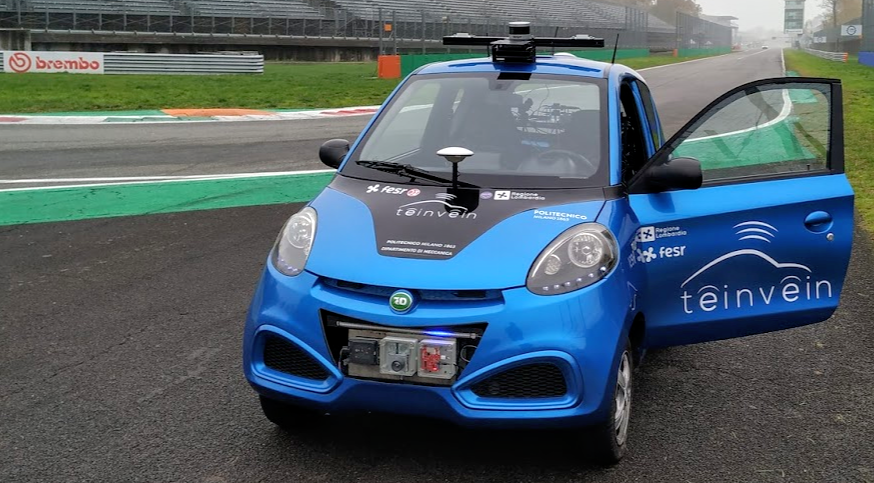}
    \caption{Inage of the experimental vehicle (Ego Vehicle) instrumented with Lidar, Radar and Camera employed for the dataset acquisition and algorithm validation}
    \label{fig:Car}
\end{figure}

Research on object tracking has led to multi-object tracking algorithms with point object representation, called point object tracking (POT). Early research with this representation merely provided information about the obstacle's kinematic state while disregarding its extent. Recent algorithms,(\cite{2020_3D_Baseline},\cite{2020_Probabilistic3D}\cite{2021_Score_Refinement}) with detect and then track approaches can also be categorized into this segment. The POT representation assumes that an object could generate at most one measurement at one time instance. One substantial drawback with this approach is the propagation of detection error to the tracking algorithm. While learning-based detectors, \cite{2019_Pointpillars} do provide sufficiently accurate detections for driving tasks, they require the use of high-resolution lidar sensors which can provide distinguishable object features. However, working with low-resolution lidar, the detections from the learning-based algorithms are not accurate due to the lack of these features. One alternative to the learning-based detections is occupancy grids-based detections \cite{mentasti2021algorithms}. The obstacles detected with this approach do not have exact object shape information and are noisy in yaw angle estimation due to the grid discretization. 
Another promising alternative is developing a tracking algorithm that eliminates the detection step or employs minimal pre-processing. 
Extended object tracking algorithms are developed with an assumption that an object can generate multiple measurements at a single time instance \cite{granstrom2017extended}. This allows for the development of tracking recursions where minimally pre-processed sensor data can be used. These measurements provide tentative information on the spatial occupancy of the object, therefore enabling the state of the obstacle to be modeled with the kinematic and extent information.\par
  
Most of the works in extended object tracking are focused on providing the state of the obstacles in cartesian coordinates in the Vehicle Reference Frame(VRF) of the ego vehicle \cite{granstrom2017extended}. While obstacle state information in cartesian coordinates is sufficient for some tasks, for applications that require information sharing, such as V2V and V2X, integration of the road model into the estimation process and state representation in curvilinear coordinates is advantageous. In particular, for information sharing, which enables easier lane level localization for participating agents~\cite{2014_Lap_Time_Curvilinear}. In addition to this, road model integration and curvilinear state representation expedite the process of dynamic path planning for the ego vehicle motion as well~\cite{2014_Improving_Reference_Trajectories}.
In literature, EOT algorithms are developed using measurements coming from standalone sensors. In~\cite{2018_Gaussian_Process_Automotiv_Radar}, authors use a high-resolution Radar sensor to develop a measurement model for a Bayesian estimation process. An extent rich and highly accurate 2D PointCloud obtained from a Lidar sensor is also employed to perform this task in~\cite{2014_GranstromEOT}. Radar measurements, which provide information on object velocity, are observed to be extremely noisy. In contrast, lidar points with highly accurate positions do not provide information on the object's speed. Fusion between these two sensors opens the door to extract a highly enriched representation of the surrounding obtained by combining complementary information. The common fusion approaches in literature can be categorized into either model based, \cite{2020_Sensor_Fusion_Multi_Modal} or data driven \cite{2021_Sensor_Fusion_Deep_learning}. \par

This work proposes an EOT algorithm that provides the obstacle state in curvilinear road coordinates by integrating a Cubic Hermite spline road model into the estimation routine. We use a fused representation of the measurement points obtained from Lidar and object detections obtained from Radar sensors using model based fusion. The proposed algorithm is validated through simulation and experimental data while also comparing with other state of the art algorithms.  

The remainder of the paper is structured in the following fashion. Related works on EOT and road model integration into tracking routine are discussed in Section \ref{StateOfArt}. In the Section \ref{SensorDataAcquisition}, the hybrid sensor fusion architecture employed in this work is presented. Section \ref{ProblemFormulation} presents the tracking problem. We present the pipeline for the EOT procedure in detail in Section \ref{Filtering_Recusrion}. Algorithm validation through simulation data and experimental data recorded by vehicle shown in Fig.~\ref{fig:Car} is presented in Section \ref{Validation}. Section \ref{Conclusion} concludes the paper.

\section{Related Works}\label{StateOfArt}

\subsection{Extended Object Tracking}
An extensive overview on EOT is presented in \cite{granstrom2017extended}. Authors have grouped EOT into three broad categories based on the different approaches used to develop measurement models. The first modeling approach assumes that measurements primarily originate from reflection points rigidly fixed to the extended objects. Authors in \cite{Radar_Resolution_Model} develop such EOT algorithm for rectangular objects using radar resolution model. This work is centered around the assumption that Radar detections from a vehicle originate from particular reflection points, for example, wheel housings, headlamps, etc.  
In the second approach, the measurement model is based on an inhomogeneous Poisson Point Process (PPP). The detections are assumed to be spread spatially around the target, and the PPP is used to model their probability. Random Matrix models are the predominant variant of this approach; in \cite{2016_GGIW_PMB_EOT}, authors develop a Gamma Gaussian Inverse Wishart Poisson Multi Bernoulli (GGIW PMB) EOT filter using this model. They assume that the measurements are Gaussian distributed around the target center while the number of measures is Poisson distributed. Based on this assumption regarding the measurement model, a Gamma-Gaussian-Inverse Wishart conjugate prior is used to develop the Bayesian recursion for filtering. In \cite{2013_EOT_Random_HyperSurface_Model}, a Random Hyper Surface Model, another variant of the spatial model, is proposed for estimating kinematic and shape estimates of ellipses and star convex shaped objects. The measurement model is developed by assuming that the measurement points are generated from sources located on the scaled boundary of the predefined shape. This scaling factor is assumed to be a random variable. Filtering recursion is performed without actually estimating the location of these assumed measurement sources. 
Finally, The third approach includes techniques that model the physics behind the sensor operation into measurement models; \cite{2016_Direct_Scattering} use direct scattering technique to perform EOT.

While some works have been developed with the predefined assumption of the object shapes, like ellipses in \cite{Probabilistic_DA_2015}, rectangles in \cite{2014_GranstromEOT}, a parametric approach that defines the boundary of an object with star convex modeling is also gaining momentum. Authors in \cite{2013_EOT_Random_HyperSurface_Model} use Fourier Coefficients to parametrize the shape of the radial function of the contour while authors in \cite{2015_Gaussian_Process} use Gaussian Processes (GP). In \cite{2015_Gaussian_Process}, authors develop an Extended Kalman Filter (EKF) based estimator by approximating Gaussian Process into a state-space model, which in turn is augmented with the kinematic state of the track. Similarly, EKF based tracker for detections obtained from Radar sensors is developed in \cite{2018_Gaussian_Process_Automotiv_Radar}. Doppler rate information obtained from the high-resolution Radar sensor is exploited to better estimate the translational and rotational velocities of the track. 

In recent developments, machine learning-based approaches are also proposed in the literature to develop measurement models. In \cite{2019_Variational_Radar_Model}, Variational Gaussian Mixture modeling is used to form a Bayesian Framework for EOT for detections obtained from Radar sensors. \cite{2021_Hierarchial_truncation_model} proposes a hierarchical truncated Gaussian model learned through raw data collected from a Radar sensor. 

\subsection{Road Model Integration into Tracking Recursion} 
A model of the geometric representation of the road can be computed online using images acquired by camera sensors \cite{2020_LaneEstimation},\cite{2015_Object_And_Situation_Assessment}. Deep learning techniques are implemented to perform image segmentation and provide road identification. However, results obtained by such online frameworks are not always robust enough to be integrated into tasks further down the pipeline due to information blackout created by occlusions or in cases of extreme weather scenarios. When the algorithm cannot rely on the information acquired through image data, localization and obstacle tracking are performed by integrating predefined HD map of the road into these routines \cite{2017_Map_Based},\cite{2017_Along_Track_Coop},\cite{2021_Integrated}. 

Integration of the road model into the tracking routine firstly provides the possibility of filtering out sensor measurements obtained from objects outside road bounds. This also enables to restrict the tracking region to be constrained within the road bounds, which is sufficient for autonomous highway applications like platooning, lane-keeping, etc. \cite{2021_Integrated}. For the development of connected architectures, authors in \cite{2017_Along_Track_Coop} demonstrate that the integration of the road model expedites the lane level localization of the participating vehicles. Furthermore, in \cite{2015_Object_And_Situation_Assessment}, authors develop a framework for lane situation assessment by integrating the road model in the point object-based estimation routine. In their work, a track-level decentralized fusion of object detections from Radar sensors is performed using the nearest neighbor approach. Track estimates provided in cartesian coordinates are later converted into road-aligned curvilinear coordinates to perform lane situation assessment using a constant curvature conversion model based on 
osculating circle assumption. The authors generate a cubic hermit spline road model from lane detections obtained from the camera sensor to perform this conversion. 
In \cite{2016_Tracking_Behaviour_Reasoning} the previous work is extended by performing fusion between Lidar, Radar, and Camera detections to generate a consistent point object measurement list. These measurements are converted into road curvilinear coordinates used to develop a tracking and behavior reasoning framework for the obstacles. This enables authors to integrate road geometry constraints into a unified interactive multi-model (IMM) tracking and behavior reasoning module.\par

Authors in \cite{2014_Improving_Reference_Trajectories} demonstrate the efficiency introduced in trajectory planning for ego vehicle motion when the road model is integrated into the planning process. Ego vehicle and road participants are localized in the road curvilinear coordinates, which makes it possible to remove the curvature effects of the road in the trajectory planning problem and hence generate requested maneuvers in the straight-road like scenarios. 

Finally, in \cite{2020_Road_Aware_Trajectory_Planning}, authors integrate road model and road constraints with learning-based techniques to perform trajectory prediction of the surrounding road vehicles. The past and present motion of the road vehicles represented as a bounding box is obtained by using  Global Navigation Satellite Systems (GNSS) and Lidar sensors. However, in a real-world driving scenario, where we assume no exchange of information between the ego vehicle and road participants, the kinematic state and extent of the obstacles can only be obtained by tracking them using exteroceptive sensors. Model Predictive Controller developed in \cite{2021_MPC_Path_Planner} assumes elliptical-shaped obstacles whose states are expressed in the curvilinear coordinates and derive the constraints for the optimization problem. The algorithm is developed assuming that the obstacle state is deterministic in the curvilinear coordinate, which does not represent the real-world driving scenario, where obstacles are generally perceived using various exteroceptive sensors. Integration of an extended object tracking algorithm in curvilinear coordinates into the pipeline provides the obstacle state for the algorithm with reasonable closeness to attainable accuracy. 

These works demonstrate the benefits of road model integration in the ego vehicle state estimation and obstacle tracking. However, sufficient attention has not been given to the accurate representation of the obstacles in the curvilinear coordinates, which is a crucial step in tasks such as road situation assessment~\cite{2015_Object_And_Situation_Assessment}, behavior prediction~\cite{2016_Tracking_Behaviour_Reasoning}, obstacle trajectory prediction~\cite{2020_Road_Aware_Trajectory_Planning}, ego vehicle trajectory generation, and controller implementation~\cite{2014_Improving_Reference_Trajectories},~\cite{2021_MPC_Path_Planner}. Similarly, the literature addresses the EOT problem for a single sensor only, either for Lidar or Radar, while no attempt has been made to exploit the fused representation for EOT. 

In this work, we propose an EOT algorithm that provides an extent rich representation of the obstacles by implementing a curvilinear coordinate-based rectangular measurement model in tracking recursion and using fused measurement points from Lidar and Radar sensors. This work presents the first approach to perform EOT in curvilinear coordinates while using fused measurements from Lidar and Radar sensors for the purpose.  
The core contributions of the paper are:
\begin{itemize}
    \item Extends the POT algorithm developed in \cite{2021_Integrated} to EOT in curvilinear road coordinates
    \item Implements the integration of road model in EOT recursion for measurement filtering and birth intensity definition
    \item Analyzes sensor fusion between lidar points and radar detections for EOT. 
    \item Demonstrates that the obstacle state representation in curvilinear coordinates increases accuracy of state estimation, especially for yaw angle estimation.
\end{itemize}

\section{Sensor Data Acquisition and PreProcessing}\label{SensorDataAcquisition}
The experimental vehicle used to validate the proposed algorithm is equipped with a Velodyne VLP-16 Lidar installed at the vehicle's roof, which provides PointCloud at $10 Hz$. In addition, two Continental ARS 408-21 Radars are installed at the front and rear of the car, and a camera is facing forward~\cite{9662926}. The detections obtained from the Radar sensors are expressed in Vehicle Reference Frame (VRF) centered at the vehicle and are obtained at $14 Hz$ . Radars already provide a cluster of detection supposedly generated from an obstacle. Hence, very limited preprocessing is required to express them in the cartesian coordinate VRF. Interested readers are referred to our previous work,\cite{2021_Integrated} for a detailed explanation of the Radar preprocessing routine.

\begin{figure}[t]
    \centering
    \includegraphics[width=\linewidth]{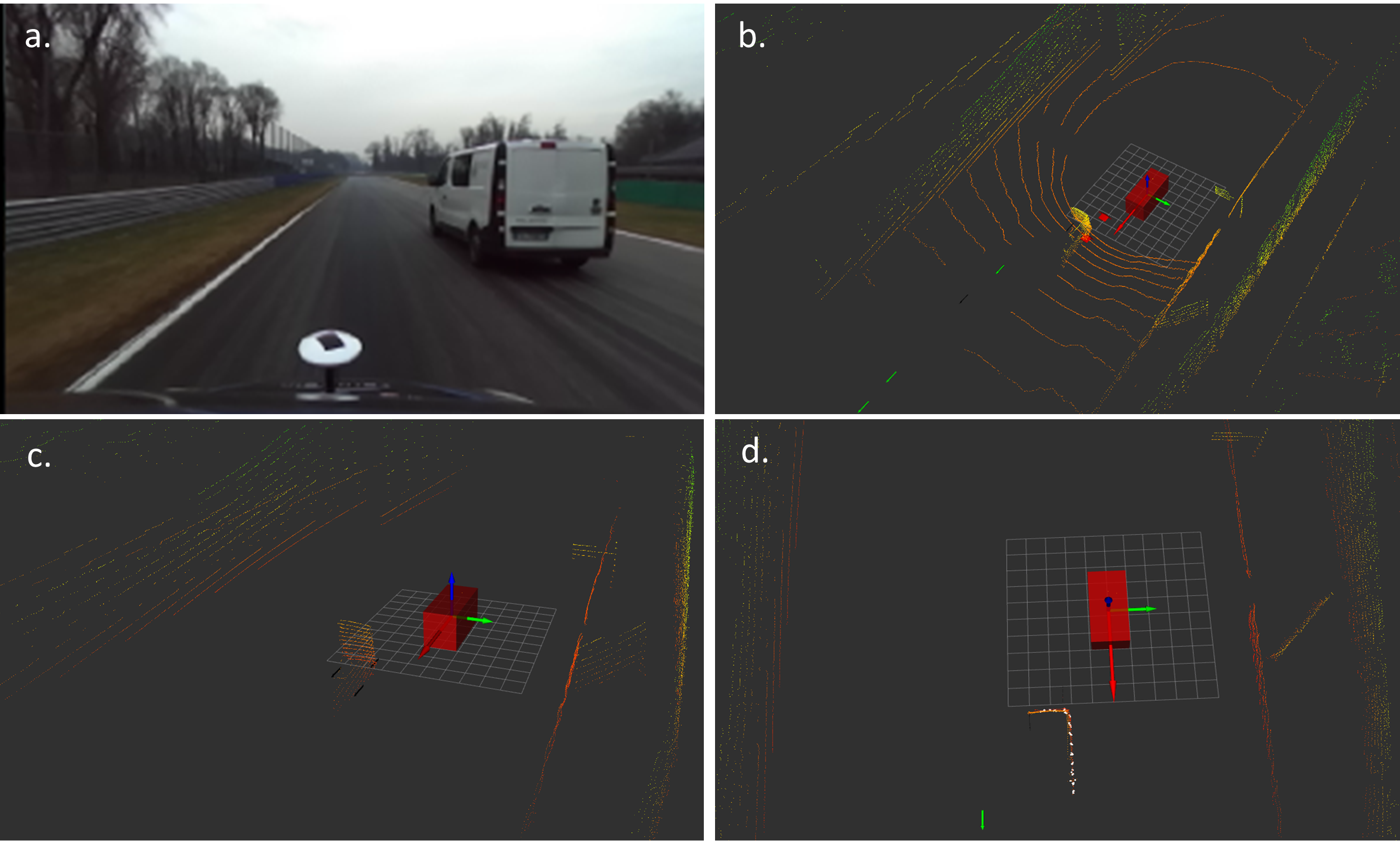}
    \caption{ Preprocessing of the Lidar PointCloud for the Experimental Scenario:  \\a. Snapshot of the  experimantal environment \\b. 3D PointCloud data obtained from the scenario, ego vehicle represented by the red bounding box \\c. filtered 3D PointCloud as points corresponding to the ground plane are removed \\d. Pointcloud from object projected to the ground plane}
    \label{fig:Lidar_Preprocess}
\end{figure}

Raw Lidar measurements are obtained as 3D PointCloud referenced to the Lidar sensor. Then, a series of preprocessing operations are performed to this 3D PointCloud data to move it to the desired 2D representation to perform EOT. Indeed, in \cite{2014_GranstromEOT}, the EOT algorithm is developed for the measurements obtained from a 2D Lidar sensor. 
Measurement in our work is also illustrated in a similar fashion, i.e, projected into ground plane, to perform rectangle fitting for developing the measurement model in the filtering recursion. The preprocessing pipeline for the Lidar data is illustrated in the Fig.~\ref{fig:Lidar_Preprocess}.
In the first step of the pipeline, the Lidar PointCloud is transformed to VRF from the Sensor Reference Frame (SRF).
Then, a plane fitting algorithm based on RANdom SAmple Consensus (RANSAC) is used to remove the points corresponding to the ground plane, as shown in~\cite{Zermas2017Fast}. Consecutively, a preliminary filtering step is implemented by exploiting the knowledge of the road bounds to remove all points outside of it. At this stage of the PointCloud preprocessing, only the points supposedly originated from the dynamic obstacles within the road bounds are left. Any obstacle within the road bound is assumed to be dynamic and is under consideration for tracking in our application. The next step involves projecting the PointCloud into the road plane using the normal direction obtained in the previous RANSAC implementation step. The output of this step is a set of observed 2D points in VRF as illustrated by  Fig. \ref{fig:Lidar_Preprocess}.d . \par
As assumed by the Poisson Point Process(PPP) spatial measurement model $p(\textbf{Z}_k|\textbf{X}_k)$ \cite{2014_GranstromEOT}, only measurements generated from the surface of the object are required. However, the output of the last preprocessing step can possibly contain measurement points corresponding to the inside of the obstacle as well. Although not visible for our experimental setup due to our tracked vehicle's nearly perfect rectangular nature, this scenario was fairly common in our simulation setup. Hence, in this step, measurements supposedly generated from the sources inside the vehicles are removed using a bearing clustering approach. In this approach, the bearing angles for all the measurement points are computed with reference to the VRF origin, and those points which fall under the threshold of some angle are clustered together. From each cluster, the closest point to the ego vehicle is chosen while all others are discarded. 

\subsection{Sensor Fusion and Measurement Clustering}
EOT algorithms are developed assuming that multiple measurement points can generate from a single object. Hence, correct clustering of these points based on their source is crucial in developing these filters. However, considering all possible clusters or partitions of these points in the update step of the filter can be computationally intractable \cite{2012_GMPHD_Tracking}. Therefore, a clustering approach is used to compute measurement clusters that are believed to stem from a single object with high confidence. Lidar is used as the primary sensor, meaning the sensor fusion module provides output only when Lidar measurements are available. This fusion architecture aims to exploit the precise object shape information obtained from the Lidar sensor and complement it with velocity information obtained from the radar sensor. Position values obtained from radar detections were observed to be extremely noisy and are hence disregarded. Meaning measurements clusters are generated only when Lidar measurements are obtained. Based on the availability of the measurement, multiple scenarios can be anticipated:

\subsubsection{Only Lidar Measurements are available}
If only Lidar measurements are available at given time instance $k$, state estimates from last time instance, $k-1$ are used to perform measurement clustering. Since the state estimates are in curvilinear coordinates ($s, n$) and measurements are in cartesian coordinates in VRF ($\textbf{E}$), a conversion is required here. First, the state positional values of the tracked objects are converted to VRF (\textbf{E}). Measurement points within the object's extent with some defined threshold are put together into a single cluster. Those points which are not selected into any cluster corresponding to the object state are further clustered using distance-based technique implemented in \cite{2012_GMPHD_Tracking}. The measurement RFS, $\textbf{Z}_k$ is represented as union of the $M_c^k$ cluster RFSs as expressed in Equation (\ref{Clustering_eq1})
\begin{equation}\label{Clustering_eq1}
    \textbf{Z}_k = \bigcup_{c=1}^{M^c_k} \textbf{C}^c_k, \textbf{ID}_k = \begin{Bmatrix} Id_{c} \end{Bmatrix}_{c=1}^{M^c_k}
\end{equation}
where, $M_c^k$ is the number of measurement clusters. In addition to this, each cluster set is assigned with an ID ,$Id_c$ to identify the nature of its origin and type of sensors used to generate it. The measurement points in cluster $\textbf{C}^c_k$ only have positional values and are expressed as:
\begin{equation} \label{Clustering_eq2}
    z_{k,j}^E=\begin{bmatrix}x_{j}^E,  y_{j}^E \end{bmatrix}_k
\end{equation}
where, $x_{j}^E$ and $y_{j}^E$ are the relative position between the ego vehicle and an obstacle.
\subsubsection{Lidar and Radar Measurements are available}
When Radar measurements are also available in addition to Lidar measurements, multi-stage clustering is employed, and the  Radar velocity components are integrated into the measurement points. First, the clustering discussed in the earlier section is performed, generating cluster sets from Lidar measurement. Then cluster to Radar detection association is performed to assign velocity values to measurement points inside the clusters. A greedy association based on the nearest-neighbor approach is done between Radar detections and cluster centroid. However, only those Lidar clusters that come within a certain distance of radar detections are assigned velocity values; the remaining ones are left unchanged. This clustering approach can provide two kinds of measurement clusters, those with measurement points having positional values only, $z_{k,j}^E=\begin{bmatrix}x_{j}^E,  y_{j}^E \end{bmatrix}_k$ and those with positional and velocity values represented in Equation (\ref{Clustering_eq3}).
\begin{equation} \label{Clustering_eq3}
    z_{k,j}^E=\begin{bmatrix}x_{j}^E,  y_{j}^E, V_{x,j}^E,  V_{y,j}^E \end{bmatrix}_k
\end{equation}
Relevant $Id$ are also assigned to these clusters based on the fusion performed. Velocities, $V_{x,j}^E$ and $V_{y,j}^E $ are computed as components of absolute velocity in $X$ and $Y$ coordinates of VRF (\textbf{E}).
 
\section{Problem Formulation}\label{ProblemFormulation}
\subsection{Measurement and Track State Representation}
The proposed algorithm aims at providing a state estimate of the obstacles around the ego vehicle in curvilinear road coordinates given the set of measurements, expressed as Random Finite Sets (RFS), $\textbf{Z}_k$. The measurement points are divided into clusters by implementing the clustering algorithm discussed in Section \ref{SensorDataAcquisition} and represented by Equation (\ref{Clustering_eq1}). Hence, the input to the filtering algorithm would be $M_k^c$ number of measurement clusters at any given time instance. 
If $N_k$ is the random unknown number of the objects of interest in the FoV of the ego vehicle at time instance $k$, the Random Finite Set (RFS) representing the object states is:
\begin{equation}\label{eqn:state_1}
    \textbf{X}_k = \begin{Bmatrix} x_{1,k},x_{2,k}...x_{N_k,k} \end{Bmatrix}
\end{equation}
where, $x_{i,k}$ is the state of the $i^{th}$ object and is made up of kinematic and shape values corresponding to the object:
\begin{equation} \label{eqn:state_2}
        x_{i,k}= \begin{bmatrix} s_i,n_i,v_i,\xi_i, \Dot{\xi}_i,L_i,W_i \end{bmatrix}^{T}_k 
\end{equation}
where, $s_i$ is the relative distance between the ego vehicle and the track $i$ in curvilinear axis, while $n_i$ is the minimum lateral distance between the road centerline and the track. $v_i$ is the track absolute velocity. $\xi_i$ and $\Dot{\xi}_i$ are the track relative yaw and yaw rate respectively to the Road Reference Frame ($\textbf{R}_o$). $L_i$ and $W_i$ are the length and width of the track. 
\begin{figure}[t]
  \begin{center}
  \includegraphics[height=5.5cm, width=9cm]{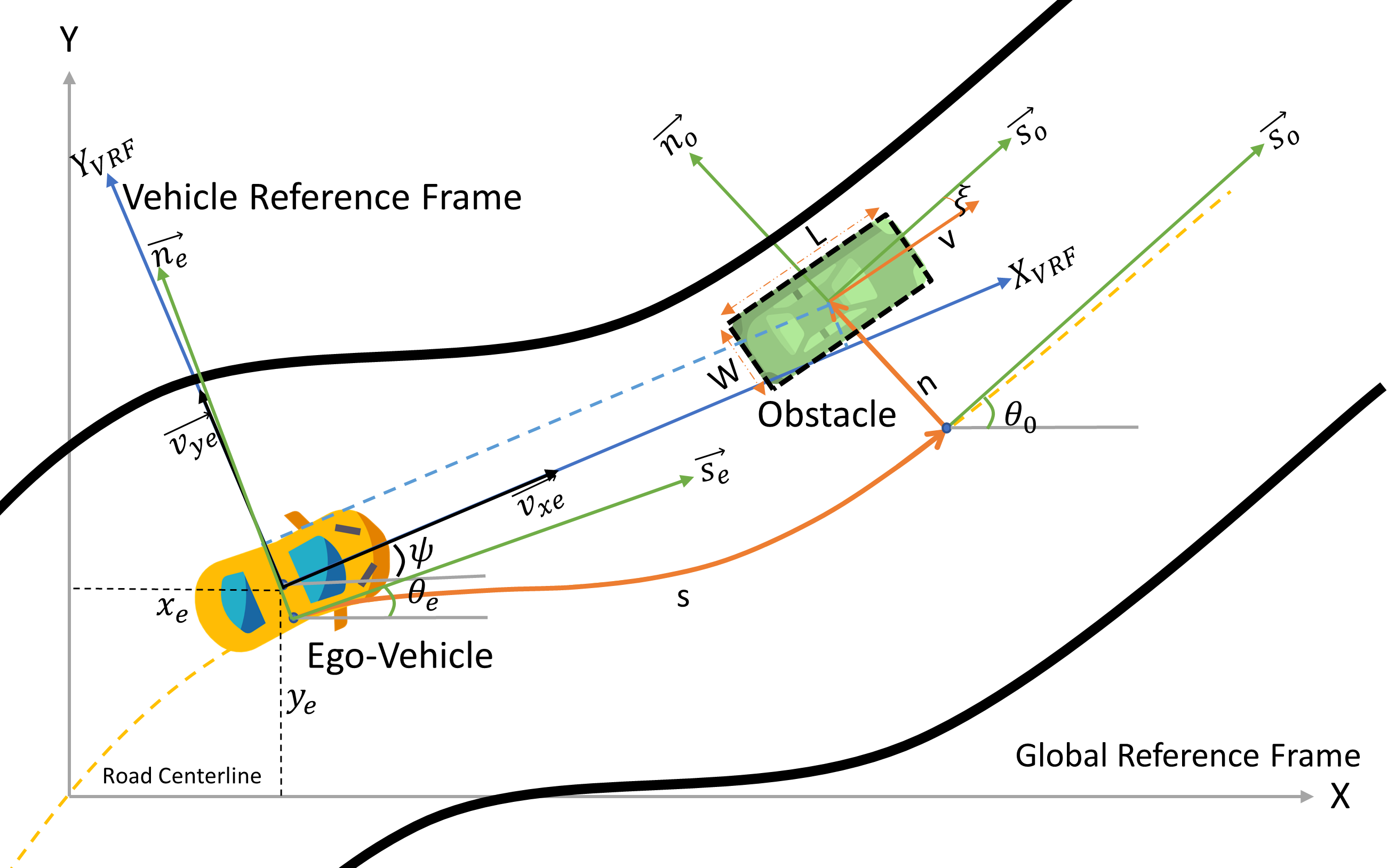}
  \end{center}
  \caption{Representation of the Ego Vehicle and Obstacle reference frames for the localization task on the road with the curvilinear road coordinates}
  \label{fig:Road1}
\end{figure}
In Fig.~\ref{fig:Road1}, $\vv{s_{e}}$ and $\vv{n_{e}}$ are abscissa and ordinate of the Road Reference Frame ($\textbf{R}_e$) centered at the nearest point corresponding to ego vehicle in road centerline. Similarly, $\vv{s_{o}}$ and $\vv{n_{o}}$ are abscissa and ordinate of the Road Reference Frame ($\textbf{R}_o$) centered at the nearest point at the road centerline corresponding to the tracked object under consideration.  

\subsection{ Multi-Object Bayesian Filtering Recursion}
A multi object Bayesian Filtering framework is developed to infer the tracked object state recursively through the posterior probability distribution function (pdf)

$p(\textbf{X}_k|\textbf{Z}_{1:k})$ given the set of measurements till time instance $k$. The filtering recursion is developed with two different steps, namely: Chapman Kolmogorov Prediction and Bayes Update. Given the prior multi-object distribution $p(\textbf{X}_{k-1}|\textbf{Z}_{1:k-1})$ and  motion model of the system $p(\textbf{X}_{k}|\textbf{X}_{k-1})$, predicted distribution $p(\textbf{X}_{k}|\textbf{Z}_{1:k-1})$ is computed using Chapman Kolmogorov Prediction formulation. To infer $\textbf{X}_{k}$, at time instance $k$, we employ a Gaussian Mean Probability Hypothesis Density (GM-PHD) filter with unscented transformations to deal with the models nonlinearities. The detailed development of the filtering recursion is discussed in the Section \ref{Filtering_Recusrion}. 

\begin{figure}[t]
  \begin{center}
  \includegraphics[height=9cm, width=9cm]{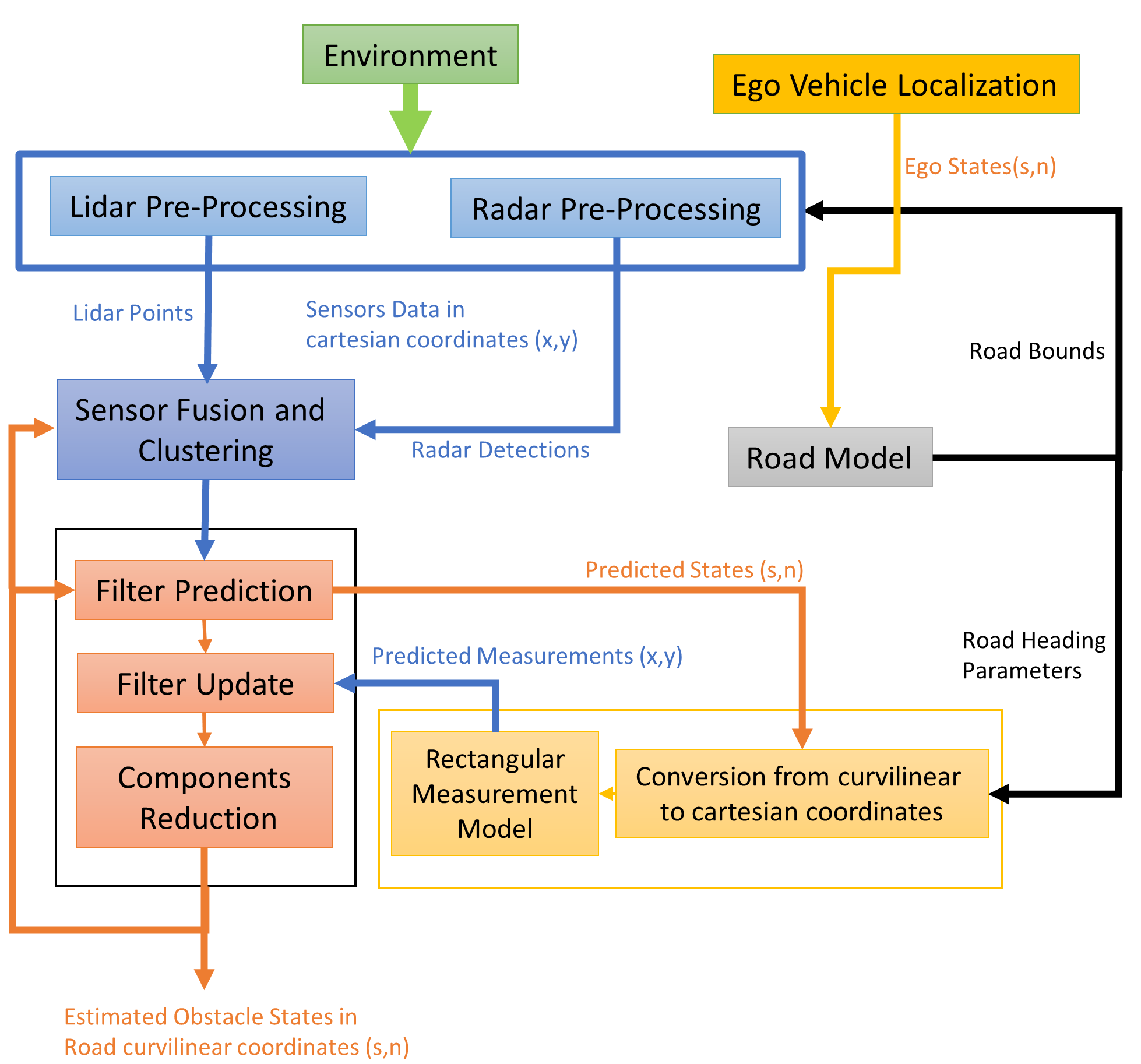}
  \end{center}
  \caption{Schema of the obstacle detection and tracking framework presented in this work. The input of the system are lidar and radar data, plus the Ego Vehicle localization. The output is a list of estimated obstacle states in curvilinear coordinates.}
  \label{fig:Framework}
\end{figure}

\section{Road model and Coordinate Conversion}
The proposed tracking algorithm is developed assuming that the road model is accurately known. A cubic hermit spline mathematical model is used to describe the road; readers are referred to \cite{2021_MPC_Path_Planner} for details on this model. For every point in the road, polynomial parameters describing the road center-line position, its heading, and curvature 30m ahead are provided. The heading angle ($\theta$) and road curvature ($\kappa$) are described using polynomial and are expressed as:
\begin{equation}\label{Road_eq1}
\begin{aligned} 
    \theta(s)=a_{\theta}s^{3}+b_{\theta}s^{2}+c_{\theta}s+d_{\theta}, \\
    \kappa(s)=a_{\kappa}s^{3}+b_{\kappa}s^{2}+c_{\kappa}s+d_{\kappa}
\end{aligned}
\end{equation}
where, $s$ is a distance between the ego vehicle and object along the road centerline, both localized in the road reference frame. The parameters of these polynomials can be accessed by localizing the ego vehicle in the global road curvilinear coordinates. The origin of this coordinate system aligns with the origin of the Global Reference Frame, which is fixed at the starting point of the experiment in the Monza Circuit.  

Coordinate conversion from curvilinear ($s,n$) to cartesian ($x,y$) coordinates, which is the first step of the proposed measurement model, can be found in our previous work, \cite{2021_Integrated}.  Pseudocode for this conversion is given in Algorithm \ref{Algorithm2}. This conversion model is based on a piece-wise Euler Integration method, which significantly differs from the model implemented in \cite{2015_Object_And_Situation_Assessment} and  \cite{2016_Tracking_Behaviour_Reasoning}. Authors in \cite{2015_Object_And_Situation_Assessment} implement a constant curvature-based conversion model, which assumes the road curvature to be constant in the conversion region. However, the proposed model depends only on the road heading angle, hence eliminating the errors propagated through the constant curvature assumption.

\section{GM-PHD Filter for Extended Object Tracking}\label{Filtering_Recusrion}
In a Probability Hypothesis Density (PHD) filter, the recursive algorithm is developed assuming the prior, $p(\textbf{X}_{k-1}|\textbf{Z}_{k-1})$, predicted $p(\textbf{X}_{k}|\textbf{Z}_{k-1})$ and posterior $p(\textbf{X}_{k}|\textbf{Z}_{k})$ distributions to be a Poisson multi-object distribution. The filtering recursion is developed by propagating the first order statistical moment of the distribution, i.e. PHD.
A PHD \textit{D(x)} when represented as weighted Gaussian Mixture leads to Gaussian Mixture Probability Hypothesis Density (GMPHD) filter and can be parameterized as:
\begin{equation} \label{PHD_eq1}
       D(x)=\sum_{i=1}^{\textit{N}} w^i \mathcal{N}(x:\mu^i,P^i)
 \end{equation}
Hence, PHD at different steps of filtering recursion is parametrized with a weight (\(w^i\)) assigned to a Gaussian Distribution, which in turn is parametrized with its mean  (\(\mu^i\)) and covariance (\(P^i\)).
The GMPHD filter discussed in \cite{2012_GMPHD_Tracking} is used for performing the EOT. Readers are referred to this work for various assumptions and derivations of the filtering algorithm.
In \cite{2014_GranstromEOT} and \cite{2012_GMPHD_Tracking}, authors use EKF as a Bayesian Estimator for performing object tracking; however, due to the highly non-linear measurement model involving multiple steps of conversion in our work, we implement UKF as our estimator. The general framework for filtering recursion in presented in the Fig. \ref{fig:Framework}.

\subsection{Filter Prediction}
The predicted RFS, $\textbf{X}_k|\textbf{Z}_{k-1}$ is computed as an union of the surviving RFS from time instance $k-1$ ,$\textbf{X}_{k-1}^S|\textbf{Z}_{k-1}$ and the new birth set at time instance $k$, $\textbf{B}_k$.
The predicted distribution in the filtering recursion remains Poisson multi-object distribution. Its PHD, $D_{k|k-1}(x)$ can be expressed as:
\begin{equation} \label{PHD_eq4}
       D_{k|k-1}(x)=D_{k|k-1}^S(x)+ \lambda_{b,k}(x)
\end{equation}
where, $\lambda_{b,k}(x)$ is intensity of birth corresponding to birth model and $D_{k|k-1}^S$ is the intensity function corresponding to the surviving objects. The birth intensity is also approximated as Gaussian Mixture. Calculation of the parameters of $D_{k|k-1}^S$ : $w_{k|k-1}^i$, $\mu_{k|k-1}^i$  and $P_{k|k-1}^i $ is based on UKF transformation through a constant turn rate motion model.
\subsubsection{Prediction of Surviving Objects} To predict the motion of the surviving objects through time instances, we implement a constant turn rate motion model in curvilinear road coordinates. Tests were carried out with constant velocity motion model as well, but we validate the algorithm with constant turn rate model to integrate possible lane change scenarios. The center of rotation of the obstacle is assumed to be in the rectangle centroid of the tracked object. The motion model is represented as:
 \begin{equation} \label{Prediction_eq1}
            \begin{bmatrix} s  \\ n \\ v \\ \xi \\ \Dot{\xi} \end{bmatrix}_{k}=\begin{bmatrix} s_{k-1}+\frac{2}{\Dot{\xi}_{k-1}}v_k\sin({\frac{\Dot{\xi}_{k-1} \delta t}{2}})\cos({\xi_{k-1}+\frac{\Dot{\xi}_{k-1} \delta t}{2}})\\ n_{k-1}+\frac{2}{\Dot{\xi}_{k-1}}v_k\sin({\frac{\Dot{\xi}_{k-1} \delta t}{2}})\sin({\xi_{k-1}+\frac{\Dot{\xi}_{k-1} \delta t}{2}})\\ v_{t-1}  \\ \xi_{k-1}+\Dot{\xi}_{k-1}\delta t \\ \Dot{\xi}_{k-1} \end{bmatrix} 
    \end{equation}
The length($L_{k-1}$) and Width($W_{k-1}$) are kept constant through prediction step. The process noise ($\omega_k$), assumed to be Gaussian and additive, is also incorporated in the motion model. The motion model can hence be represented by the Equation  (\ref{Prediction_eq2}).
\begin{equation}\label{Prediction_eq2}
         x_k=f(x_{k-1})+g(\omega_k)
\end{equation} 
where, $f(x_{k-1})$ represents the motion model given by Equation (\ref{Prediction_eq1}) and $g(\omega_k)$ process noise. 
\begin{figure}[t]
        \centering
        \includegraphics[width=\linewidth]{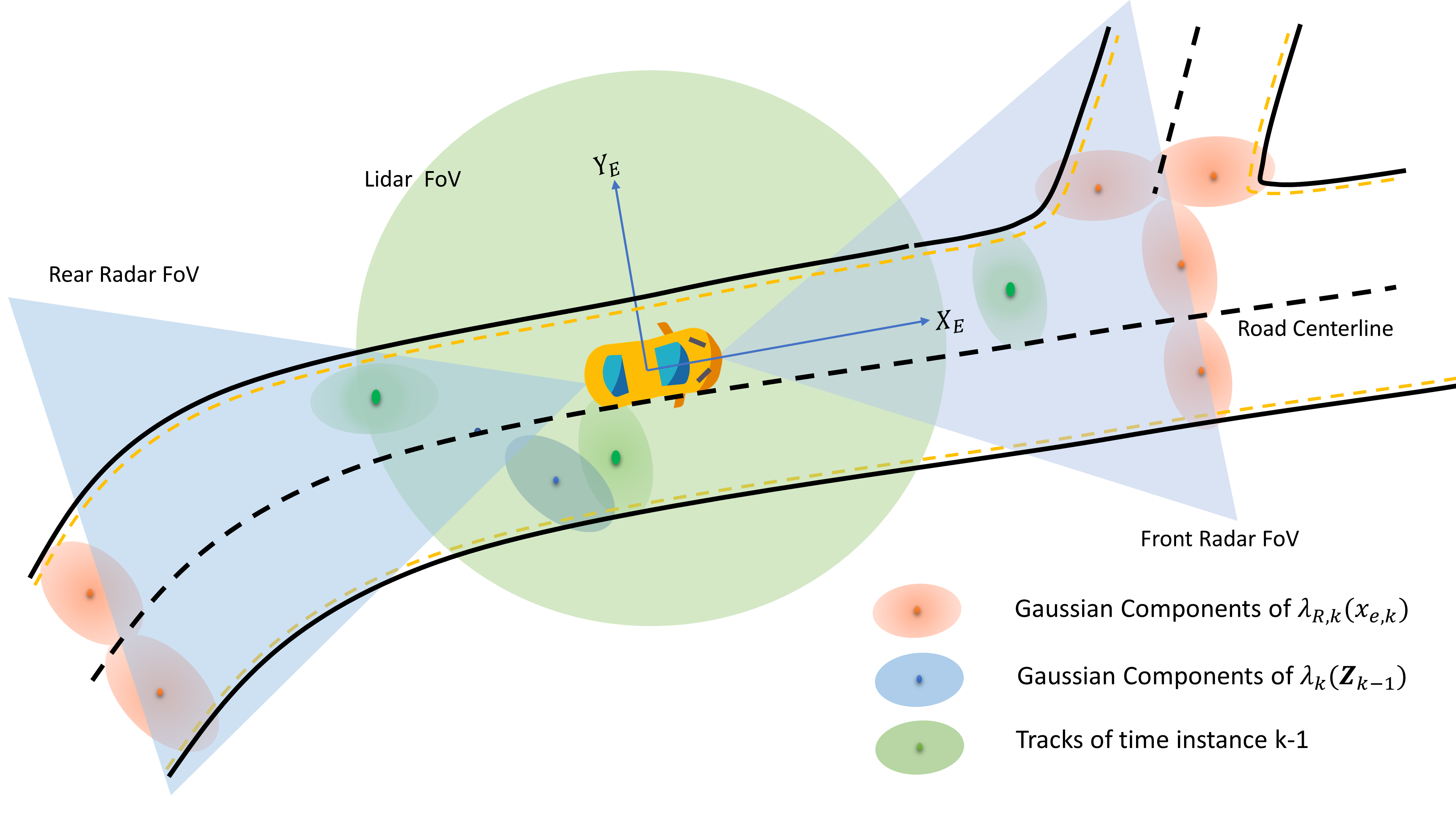}
        \caption{Birth Components based on sensor FoV and Road Geometry and measurements of previous time instance.}
        \label{fig:Birth1}
\end{figure}
$\mu_{k|k-1}^i$  and $P_{k|k-1}^i $, which are the mean and covariance of the weighted Gaussian components are computed using the unscented transformation. Furthermore, the weight of each Gaussian component,  $w_{k|k-1}^i$  is scaled by the Probability of Survival $P_S$, which  is assumed to be constant and state independent value in our work.

\subsubsection{Addition of Birth Components} We exploit information of road network and bounds to define the birth intensity, $\lambda_{b,k}(x)$. In addition to this, those measurement points which are very far from the preexisting tracks in terms of Mahalanobis Distance are also used to initialize new births components. Fig.~\ref{fig:Birth1} illustrates the Gaussian components used to define these birth process. 
\begin{equation} \label{Prediction_eq3}
        \lambda_{b,k}(x)=\lambda_{R,k}(x_{e,k})+\lambda_k(\textbf{Z}_{k-1})
\end{equation}
Ego vehicle localization in the road, $x_{e,k}$, the road geometry and the FoV definition of sensors are used to define the road based Gaussian components, $\lambda_{R,k}(x_{e,k})$. Disregarded measurement clusters from last time instance $\textbf{Z}_{k-1}$ are used to define intensity corresponding to measurements, $\lambda_k(\textbf{Z}_{k-1})$. State representation in curvilinear road coordinates further facilitates this definition of the birth components, for example, when the road bifurcates in the scenario illustrated in Fig.~\ref{fig:Birth1}.

\subsection{Filter Update}
Prediction step outputs $p(\textbf{X}_k|\textbf{Z}_{k-1})$, which is multi-object Poisson distribution of Poisson RFS $\textbf{X}_k|\textbf{Z}_{k-1}$. The PHD of this distribution, calculated by using equation (\ref{PHD_eq4}) is also approximated as a Gaussian mixture. A measurement model is developed to update these intensity parameters, which outputs predicted measurement points in cartesian coordinates of VRF from the state in curvilinear road coordinate. 
\subsubsection{Measurement Model}A single object measurement model is discussed in this section, which is later integrated into the tracking iteration for the multi-object tracking process. Let $\textbf{C}_k = \begin{Bmatrix} z_l\end{Bmatrix}_{l=1}^{|\textbf{C}_k|}$  be the set of measurements generated from an object $x_{i,k}$ clustered together. A  measurement likelihood $p(\textbf{C}_k|x_{i,k})$ needs to be calculated to define the measurement update step.  Measurement points within this cluster are assumed to be independent of each other, hence the measurement likelihood can be computed as:
\begin{equation} \label{MeasModel_eq1}
    p(\textbf{C}_k|x_{i,k}) = \prod_{l=1}^{|\textbf{C}_k|} p(z_l|x_{i,k})
\end{equation}
A measurement point $z_l$ can be assumed to be generated from a measurement generating point $y_l$ with some noise $e$, i.e. $z_l = y_l + e$. Removing the point indices for clearer representation, the likelihood of point $z$ is expressed as: 
\begin{equation} \label{MeasModel_eq2}
    p(z|x) = \int p(z|y)p(y|x)dy
\end{equation}
and approximated by Gaussian mixture as:
\begin{equation}\label{MeasModel_eq3}
    p(z|x) = \sum_{i=1}^N w^i \mathcal{N}(z:y^i(x),R^i) , \sum _{i=1}^N w^i = 1
\end{equation}\par
where, N is the number of Gaussian components used to approximate the likelihood. This value is equal to the numbers of measurement points in the given cluster, $M_k^c$ under consideration for track state update. 
It is also assumed that each measurement generating point, $y$ generates a single measurement point only. Development of this measurement model requires the computation of the measurement generating points $\textbf{Y}_k$ from the known object state $x_{i,k}$ and obtained measurement cluster $\textbf{C}_k$. Since the object state is in curvilinear coordinate($s,n$), the first step of this process would be to compute the state in cartesian coordinate ($x,y$) in VRF ($\textbf{E}$).

\begin{figure}[t]
        \begin{center}
        \includegraphics[width=8cm]{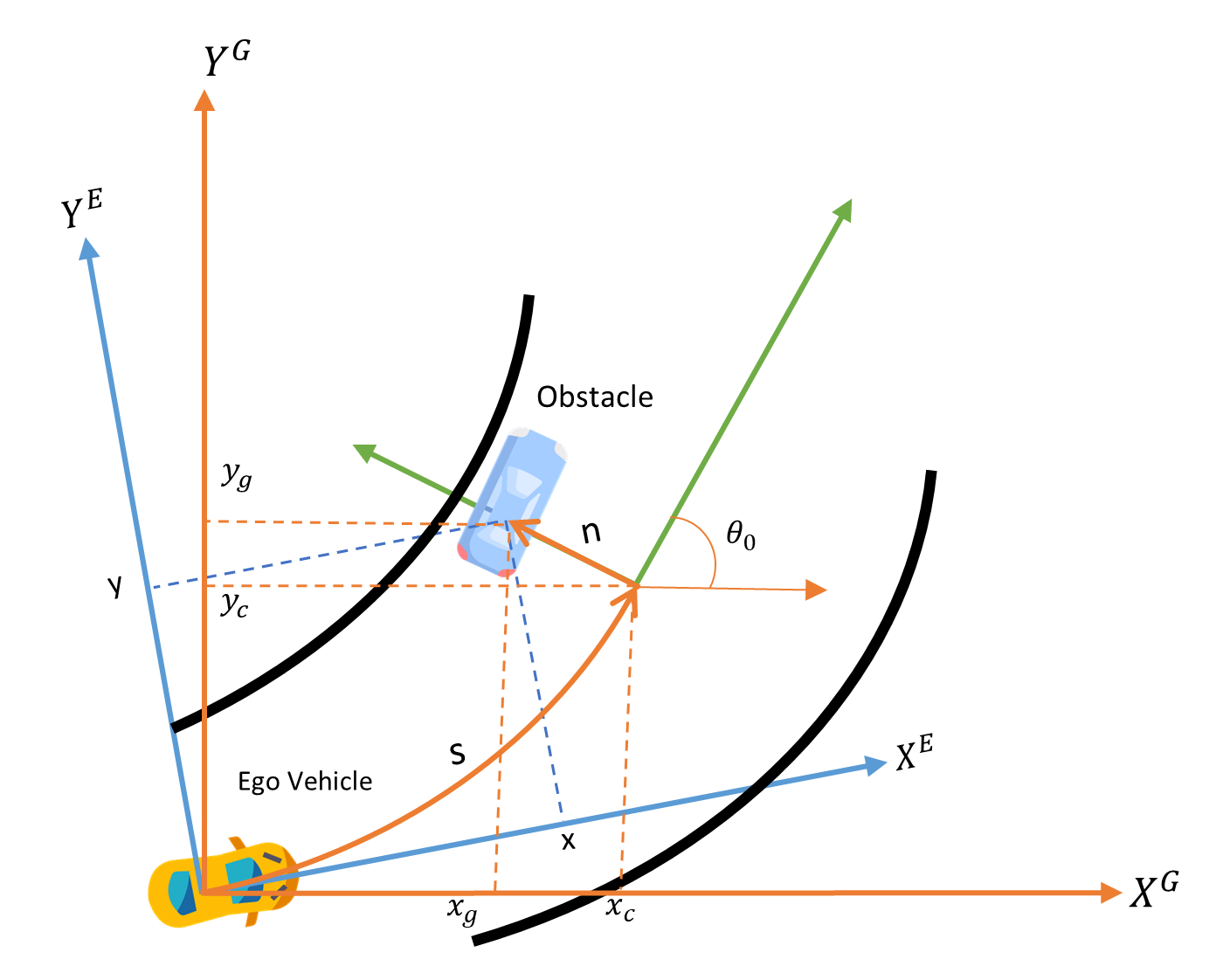}
        \end{center}
        \caption{Representation of the curvilinear and cartesian coordinate system in Vehicle Reference Frame}
        \label{fig:Euler}
\end{figure}

\textbf{Coordinate Conversion:} This conversion process is illustrated in the Fig.~ \ref{fig:Euler} and pseudocode in Table \ref{Algorithm2}. The pseudocode algorithm provides the conversion for the positional values of the state( $s,n$ to $x,y$), the velocity value remains the same while the yaw and yaw rate conversions are given by equations (\ref{MeasModel_eq4}).  

\begin{algorithm}
\caption{Curvilinear To Cartesian :Euler Model}
\label{Algorithm2}
\begin{algorithmic}
    \STATE \textbf{Input:} (x, $\theta_r$, $x_e$ ,\textit{EgoPosition})
    \STATE \textbf{Computations:}
    \STATE $\Bar{\theta}=\theta_r(EgoPosition)$ 
    \STATE \texttt{Discritize} $s_a \longleftarrow 0:\delta s:s$
    \STATE $N \longleftarrow \texttt{length}(s_a)$
    \STATE $\theta_a \longleftarrow \Bar{\theta}(1)s_a^3+\Bar{\theta}(2)s_a^2+\Bar{\theta}(3)s_a+\Bar{\theta}(4)$
    \STATE $\theta_o \longleftarrow \theta_a(end)$
    \STATE \texttt{Position Coordinate Conversion}
    \FOR{\texttt{n=2 to N}}
        \STATE $x^n \longleftarrow x^{n-1}+\delta s \cos(\theta_a(n))$
        \STATE $y^n \longleftarrow y^{n-1}+\delta s \sin(\theta_a(n))$
    \ENDFOR
    \STATE $x_c \longleftarrow x^N$
    \STATE $y_c \longleftarrow y^N$
    \STATE $ x_g \longleftarrow x_c+n\cos(\frac{\pi}{2}-\theta_0)$
    \STATE $y_g \longleftarrow y_c+n\sin(\frac{\pi}{2}-\theta_0)$
    \STATE $\xi \longleftarrow x_e(2)$
    \STATE $\psi \longleftarrow x_e(3)$ $\longleftarrow \text{Ego Vehicle Yaw Angle}$
    \STATE $x \longleftarrow \cos(\psi)x_s-\sin(\psi) y_s$
    \STATE $y \longleftarrow \sin(\psi)x_s+\cos(\psi) y_s+x_e(1)\cos(\xi)$
    \STATE \textbf{Output:} (x,y)
\end{algorithmic}
\end{algorithm}

\begin{equation} \label{MeasModel_eq4}
\begin{aligned}
      \psi = \xi + \theta_o -\psi_e\\
      \Dot{\psi} = \Dot{\xi} + \kappa_o. v. cos(\xi) -\Dot{\psi}_e
\end{aligned}
\end{equation}
where, $\kappa_o$ is the curvature of the road component corresponding to the obstacle position. $\psi_e$ and $\Dot{\psi}_e$ are yaw and yaw rate of ego vehicle. The output of this conversion step is the object state in VRF (\textbf{E}), $x_{i,k} = \begin{bmatrix} x ,y ,v,\psi ,\Dot{\psi},L,W\end{bmatrix}^T_{i,k}$. It is important to notice that the yaw and yaw rate are relative values in VRF. \par
\textbf{Computing Measurement Generating Points: }In the second stage, the measurement generating points are computed from the object state by closely following the model developed by \cite{2014_GranstromEOT} with modifications to include velocity components. The process is discussed here for completeness. In \cite{2014_GranstromEOT}, authors only use measurements obtained from the Lidar sensor to develop the measurement model, however, this work further extends it by using the measurement points obtained from sensor fusion module from Lidar and Radar sensors. To develop this model, the Lidar sensor is taken as a primary sensor, i.e., measurement points are used to update the object state only when measurements from Lidar are available. The measurement points in the cluster $\textbf{C}_k$ are sorted in counter-clockwise direction based on their bearing angle, and predicted measurement generating points are similarly arranged, eliminating the need to perform the association between these points. The likelihood then becomes, 
\begin{equation}\label{MeasModel_eq5}
p(\textbf{C}|x) = \prod_{l=1}^{|C|} \mathcal{N}(z_l; y_l(x),\sigma^2_r I_2)= \mathcal{N}(z_C;y_C(x),S_C)
\end{equation}
where, $z_C$ is measurement points vertically concatenated and  $y_C(x)$ is predicted measurement generating points vertically concatenated as well. $S_C$ is the measurement innovation matrix.  The model is developed with an assumption that the sensor can see a maximum of two sides of the rectangular object at a time instance. Based on this assumption, two models are developed.

\textit{Single Sided Measurements:}
For the state $x_i$, let $\alpha_1$,$\alpha_2$,$\alpha_3$ and $\alpha_4$ be the angles made by normal of the sides of the rectangle. Let $\beta$ represent the angle made by the vector from the first to last measurement point in the cluster. The measurements are associated to $i^{th}$ side of the rectangle, which satisfy the condition:
\begin{equation}
    i_{min} = argmin_i|\alpha_i-\beta+\pi/2|
\end{equation}
The measurement generating points are computed to be spread along the associated side based on the extent of the spread of the measurement points in $\textbf{C}_k$. \par
\textit{Double Sided Measurements:} If the measurements in $\textbf{C}_k$ are deemed to be two sided measurements, a corner index is computed to separate the measurements into two single sided measurements. The corner index, $n$ is computed by least squares fitting lines. By doing so, $\textbf{C}_k$ is split into two sub sets, $\textbf{C}_{k,1} = \begin{Bmatrix} z _l\end{Bmatrix}_{l=1}^n$ and $\textbf{C}_{k,2} = \begin{Bmatrix} z _l\end{Bmatrix}_{l=n+1}^{|\textbf{C}_k|}$. Now, the single sided model is applied to each sub set to obtain the predicted measurement generating points. 
\subsubsection{State Update}
The posterior PHD is computed as:
\begin{equation}
    D_{k|k}(x) = D_{k|k}^{ND}(x)+\sum_{\textbf{C}_k\in \textbf{Z}_k} D_{k|k}^D(x,\textbf{C}_k)
\end{equation}
where, the PHD, $D_{k|k}^{ND}(x)$ is computed for the undetected objects, while $D_{k|k}^D(x,\textbf{C}_k)$ is computed for detected objects using the measurement clusters which are the output of the sensor fusion module. The PHD corresponding to the detected objects is given by:
\begin{equation} \label{Update_eq1}
       D_{k|k}^D(x,\textbf{C}_k)=\sum_{i=1}^{\textit{N}_{k|k-1}} w_{k|k}^{i,\textbf{C}_k}\mathcal{N}(x:\mu_{k|k}^{i,\textbf{C}_k},P_{k|k}^{i,\textbf{C}_k})
\end{equation}
To compute the posterior parameters, $w_{k|k}^i$, $\mu_{k|k}^i$ and $P_{k|k}^i$ for the cluster $\textbf{C}_k$, update recursion of UKF is used. What follows is the update of the Gaussian components of the predicted density with the a measurement cluster $\textbf{C}_k$; it should be noted that the measurement points at time instance $k$, $\textbf{Z}_k$ would be divided into multiple cluster as an output of the clustering algorithm discussed in Section \ref{SensorDataAcquisition}.

The predicted measurement mean (${y}_{k|k-1}^{i,C}$) and Innovation covarience matrix ($S_{k|k-1}^{i,C}$) for Gaussian component $i$ and cluster $\textbf{C}_k$ can be calculated by using the Equations (\ref{Update_eq2}) and (\ref{Update_eq3}).
\begin{equation}\label{Update_eq2}
    {y}_{k|k-1}^{i,C}=\sum_{\alpha=1}^{2n+1}w_{\alpha}\zeta_{\alpha,k|k-1}^{i,C}
\end{equation} 
\begin{equation}\label{Update_eq3}
    S_{k|k-1}^{i,C}=\sum_{\alpha=1}^{2n+1}w_{\alpha}[\zeta_{\alpha,k|k-1}^{i,C}-{y}_{k|k-1}^{i,C}][\zeta_{\alpha,k|k-1}^{i,C}-{y}_{k|k-1}^{i,C}]^{T}+R_{C}
\end{equation}
where, $R_C$ is the measurement noise covarience matrix obtained as $R_C = blkdiag(R_k,R_k......R_k)$. Definition of $R_k$ depends on the type of the measurement cluster $\textbf{C}_k$, whether it has standalone Lidar points or the fused points form Lidar and Radar sensors. The noise for Lidar only points consider the noise between the measurement generating points and the measurement in spatial space while fused ones also consider the radar noise for the velocity components.  $\zeta_{\alpha,k|k-1}^{i,C}$ are the predicted measurement sigma points computed by passing state sigma points,$\chi_{\alpha,k|k-1}^{i}$ through the measurement model discussed earlier.  
Finally, the cross correlation matrix between the predicted state and predicted measurement is calculated in Equation (\ref{Update_eq4}).
\begin{equation}\label{Update_eq4}
    T_{k|k-1}^{i,C}=\sum_{\alpha=1}^{2n+1}w_{\alpha}[\chi_{\alpha,k|k-1}^{i}-{\mu}_{k|k-1}^{i}][\zeta_{\alpha,k|k-1}^{i,C}-{z}_{k|k-1}^{i,C}]^{T}
\end{equation}
Next, the Kalman gain can be calculated using this cross correlation matrix as:
\begin{equation}\label{Update_eq5}
    K_{k|k-1}^{i,C}=T_{k|k-1}^{i,C}(S_{k|k-1}^{i,C})^{-1}
\end{equation}
and the predicted mean of Gaussian component $i$ is updated by measurement cluster $\textbf{C}_k$ as:
\begin{equation}\label{Update_eq6}
    \mu_{k|k}^{i,C}=\mu_{k|k-1}^{i}+K_{k|k-1}^{i,C}(z_{k}^C-y_{k|k-1}^{i,C})
\end{equation}
while, the covarience matrix is updated as:
 \begin{equation}\label{Update_eq7}
    P_{k|k}^{i,C}=P_{k|k-1}^i-(K_{k|k-1}^{i,C})S_{k|k-1}^{i,C} (K_{k|k-1}^{i,C})^{T}
\end{equation}
the weight of the each Gaussian component is also updated as:
\begin{equation} \label{Update_eq8}
    {w}_{k|k}^{i,C}=\frac{P^D w_{k|k-1}^i \Gamma^i L^{i,C}}{\delta_{|C|,1}+\sum_{i=1}^{\textit{N}_{k|k-1}} \Gamma^i P^D w_{k|k-1}^{i} L^{i,C}}
\end{equation}
where,$L^{i,C}$ is computed using Equation (\ref{Update_eq9}): 
\begin{equation}\label{Update_eq9}
    L^{i,C}= \mathcal{N}(z_k^C:y_{k|k-1}^{i,C},S_{k|k-1}^{i,C}) \prod _{z_k \in \textbf{C}_k} \frac{1}{\lambda_k c_k(z_k)}
\end{equation}
and $\Gamma^i$ is computed as:
\begin{equation}\label{Update_eq10}
    \Gamma^i = e^{-\gamma^i}(\gamma^i)^{|C|}
\end{equation}
Standard mixture reduction techniques like: pruning, merging and capping are employed to ensure the tractability of the algorithm.

\section{Algorithm Validation}\label{Validation}

\begin{figure}
    \centering
    \includegraphics[width = 0.5\textwidth]{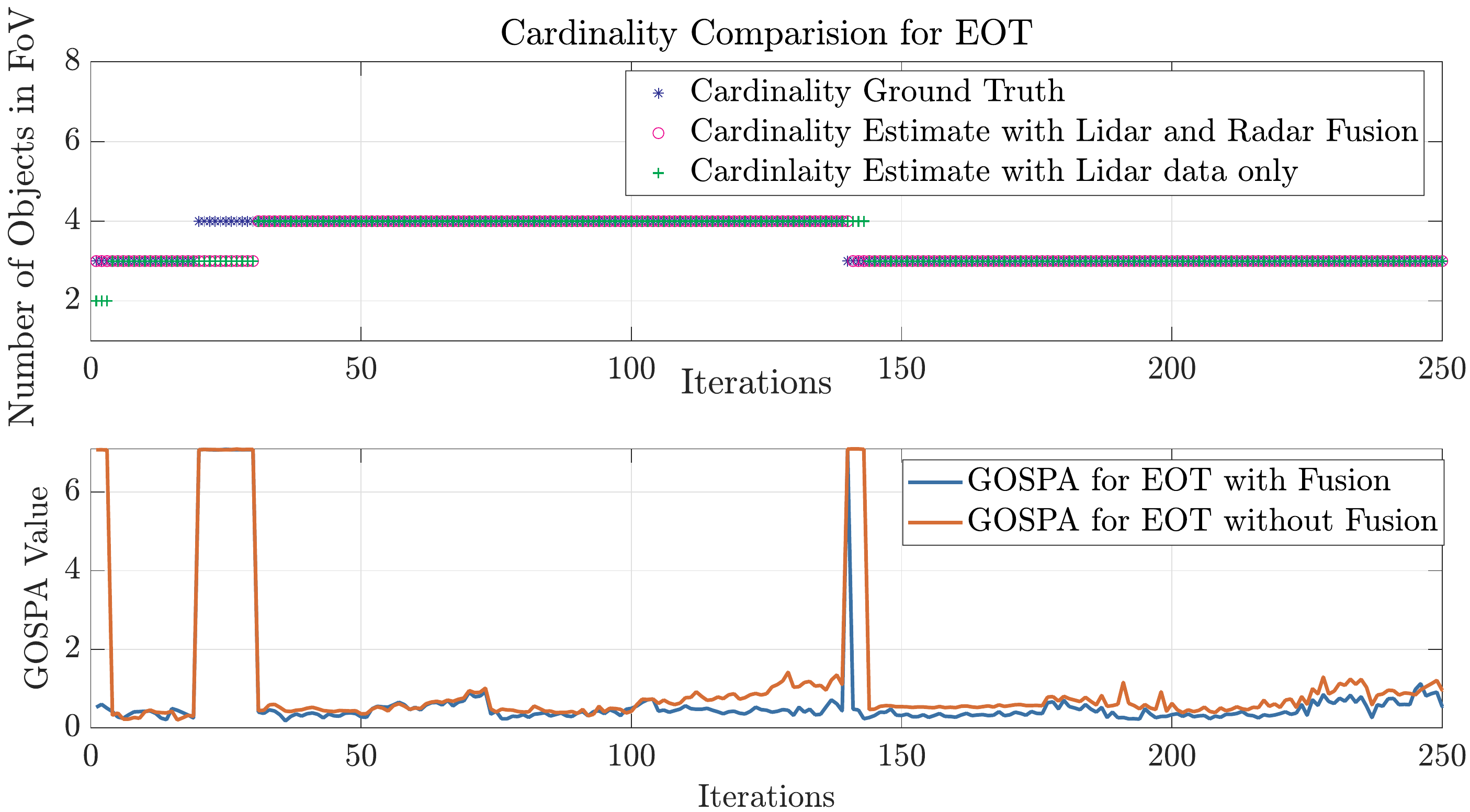}
    \caption{Cardinality and GOSPA values computed for estimated states for multi-object tracking scenario}
    \label{fig:SimulationResults}
\end{figure}

The proposed EOT algorithm provides estimates of the kinematic and extent state of objects within road bounds in curvilinear road coordinates. The state estimates are computed from the measurements obtained from Lidar and Radar sensors. Precomputed HD road map is integrated into the EOT GM-PHD filtering routine. The algorithm is validated through simulation and experimental data. 
\subsection{Simulation Validation}
A multi-object scenario is developed in Matlab Driving Scenario Designer application to validate the algorithm. Here, we present one of the scenarios with high road curvature and obstacles performing various maneuvers. The simulated scenario is made available in GIF format \href{https://polimi365-my.sharepoint.com/:i:/g/personal/10622973_polimi_it/EVj0bfFOQvpFqIofVhXeLu4BV6NzxUGPujSEcUyFTlUNWg?e=UzB4Pi}{here}. This visualization contains all the sensor data along with the obtained state estimate in curvilinear coordinate. 

Fig. \ref{fig:SimulationResults} illustrates how the algorithm can consistently provide obstacle state estimates with reasonable accuracy. The scenario is simulated with two different setups, one using both Lidar and Radar sensor while another with Lidar data only. Results in Fig. \ref{fig:SimulationResults} show that EOT employing sensor fusion between lidar and radar sensor performs better in relation to the localization error component of the Generalized Optimal sub-pattern assignment (GOSPA) metric, \cite{2017_GOSPA}. The peaks in the GOSPA values are observed due to occlusion in the driving scenario and error in ground truth cardinality estimation for defined FoV in the simulation environment.  \par
\begin{figure}
    \centering
    \includegraphics[width = 0.5\textwidth]{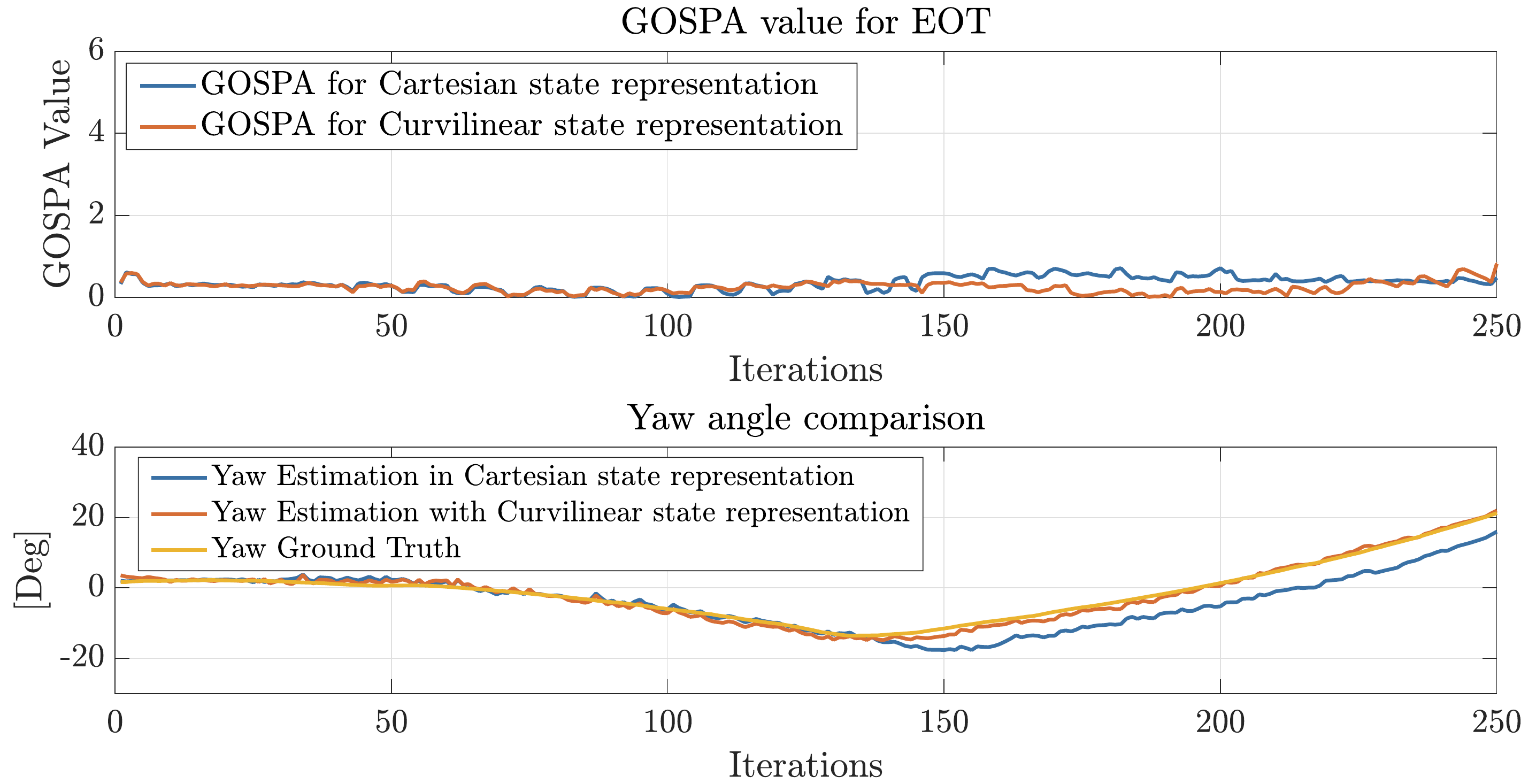}
    \caption{Top: GOSPA values computed with state representation in cartesian and curvilinear coordinates Bottom: Yaw angles estimates obtained with state representation in cartesian and curvilinear coordinates.}
    \label{fig:SimulationResults1}
\end{figure}

Furthermore, we observed that the obstacle state representation in curvilinear coordinates enables for better accuracy in obstacle state estimation. In Fig. \ref{fig:SimulationResults1}, the GOSPA values are computed for two different approaches for single object tracking scenario, in first approach, the obstacles states are represented in cartesian coordinates while in second approach states are represented in the curvilinear coordinates. For comparison with the ground truth, the states computed in curvilinear coordinates are converted to cartesian coordinates. It can be observed that the state estimation in curvilinear coordinates allows for better accuracy in terms of object localization and especially computation of the yaw angle. (GIF for the scenario available \href{https://polimi365-my.sharepoint.com/:i:/g/personal/10622973_polimi_it/EQHE5QNA1lFMpeKlIoiXD4UBlXzUYzhvFqbZng05Yt5kMg?e=zOOXDc}{here}. ).

\subsection{Experimantal Validation}
\begin{figure}
    \centering
    \includegraphics[width =9cm]{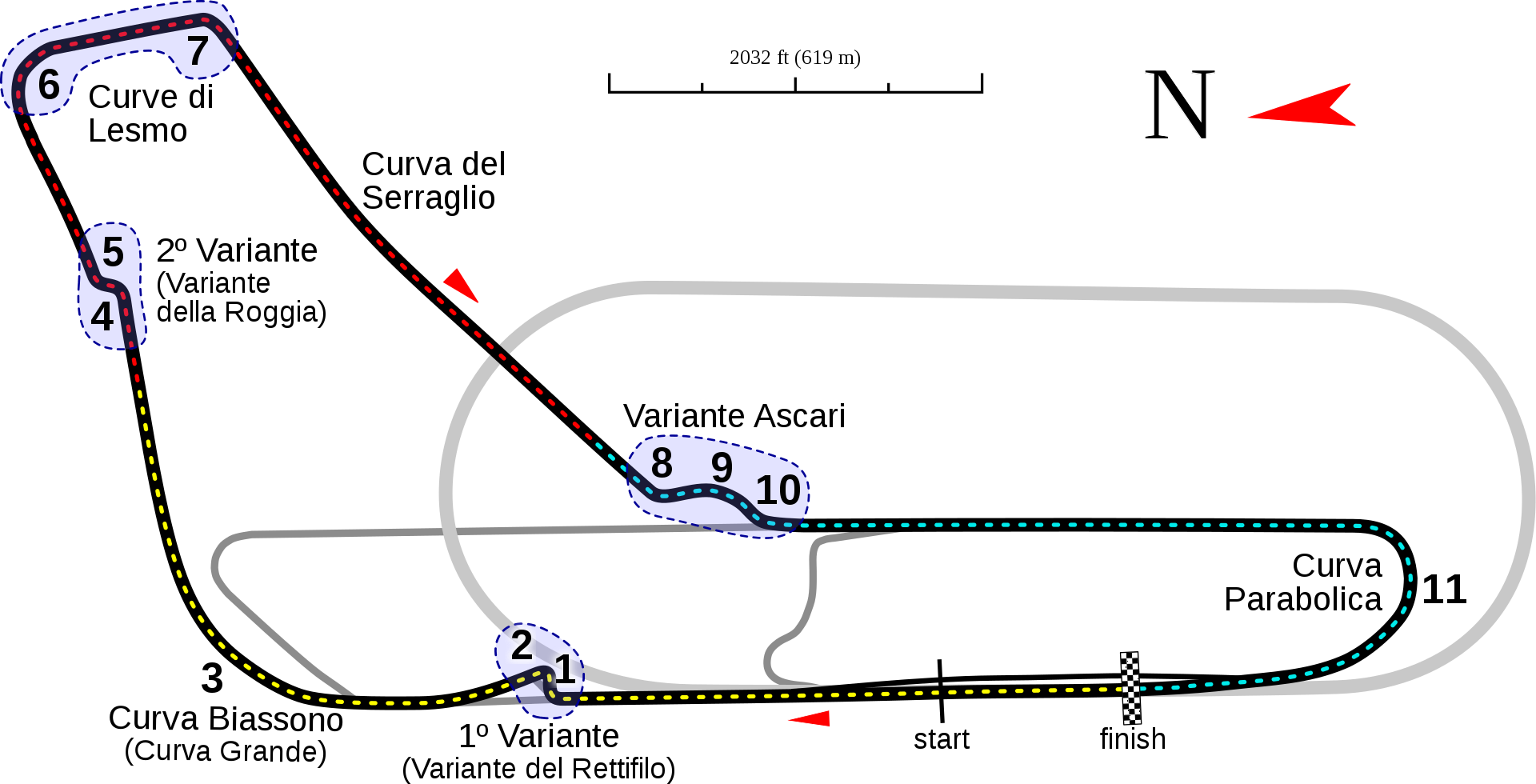}
    \caption{ Experimental Test site, Monza Circuit}
    \label{fig:Monza}
\end{figure}

For safety reasons and to acquire an accurate ground truth, the algorithm is validated using data collected during experimental campaigns at Monza Eni Circuit, as shown in Fig.~\ref{fig:Monza}.
The experiment consists of the ego prototype and another tracked vehicle installed with RTK corrected GPS (Real Time Kinematic Global Positional System). The tracked vehicle used in the experiments is FIAT Talento, with size $4.8m$ ×
$2m$ × $2m$. The algorithm is analyzed by dividing a single lap run at the Monza circuit into various segments based on mutual maneuvers between the ego vehicle and the obstacle and the degree of the road curvature. 
Given that the output of the Lidar preprocessing is at $10Hz$, the object state estimation routine runs at $10Hz$ on a soft real-time-based Robot Operating System (ROS) system~\cite{quigley2009ros}.

\begin{figure*}[t]
\centering
\subfloat[Comparison of longitudinal state estimates ]{\includegraphics[width=0.5\linewidth]{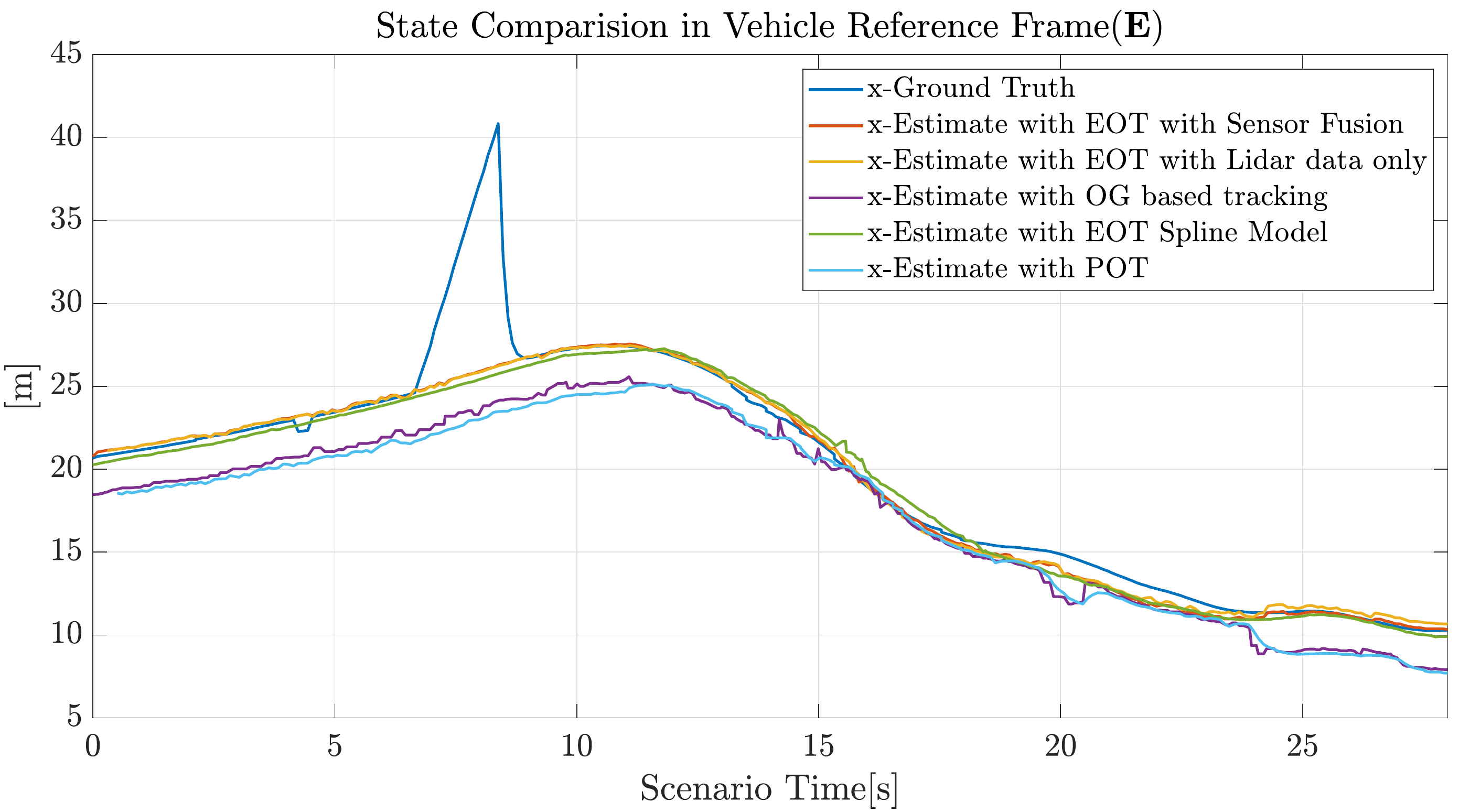}%
\label{fig:x_comp}}
\hfil
\subfloat[Comparison of lateral state estimates ]{\includegraphics[width=0.5\linewidth]{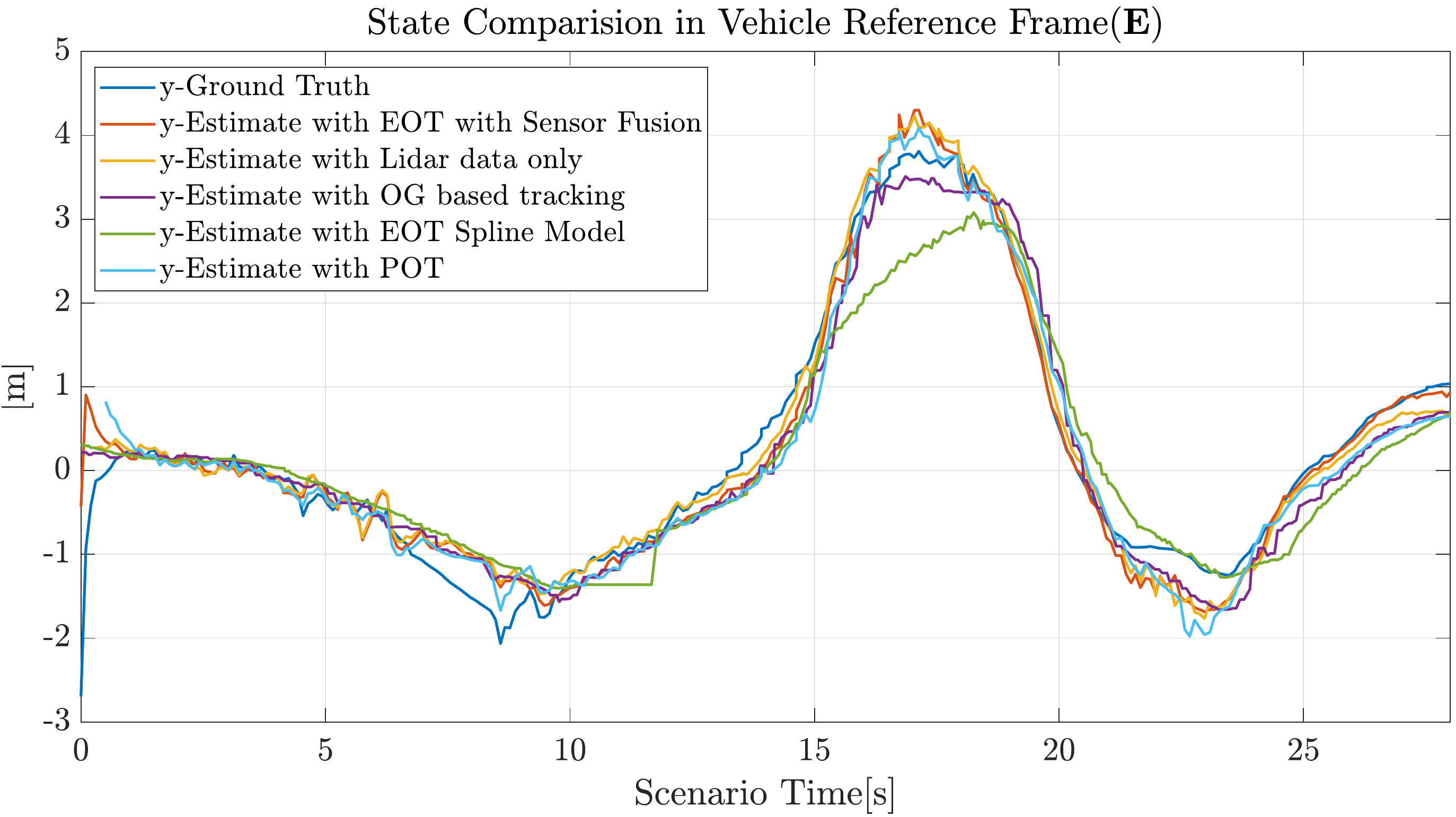}%
\label{fig:y_comp}}
\caption{Positional state estimates obtained from different tracking approaches with ground truth for Scenario 1 }
\label{fig:position_comp}
\end{figure*}

\begin{figure*}[t]
\centering
\subfloat[Comparison of yaw angle estimates ]{\includegraphics[width=0.5\linewidth]{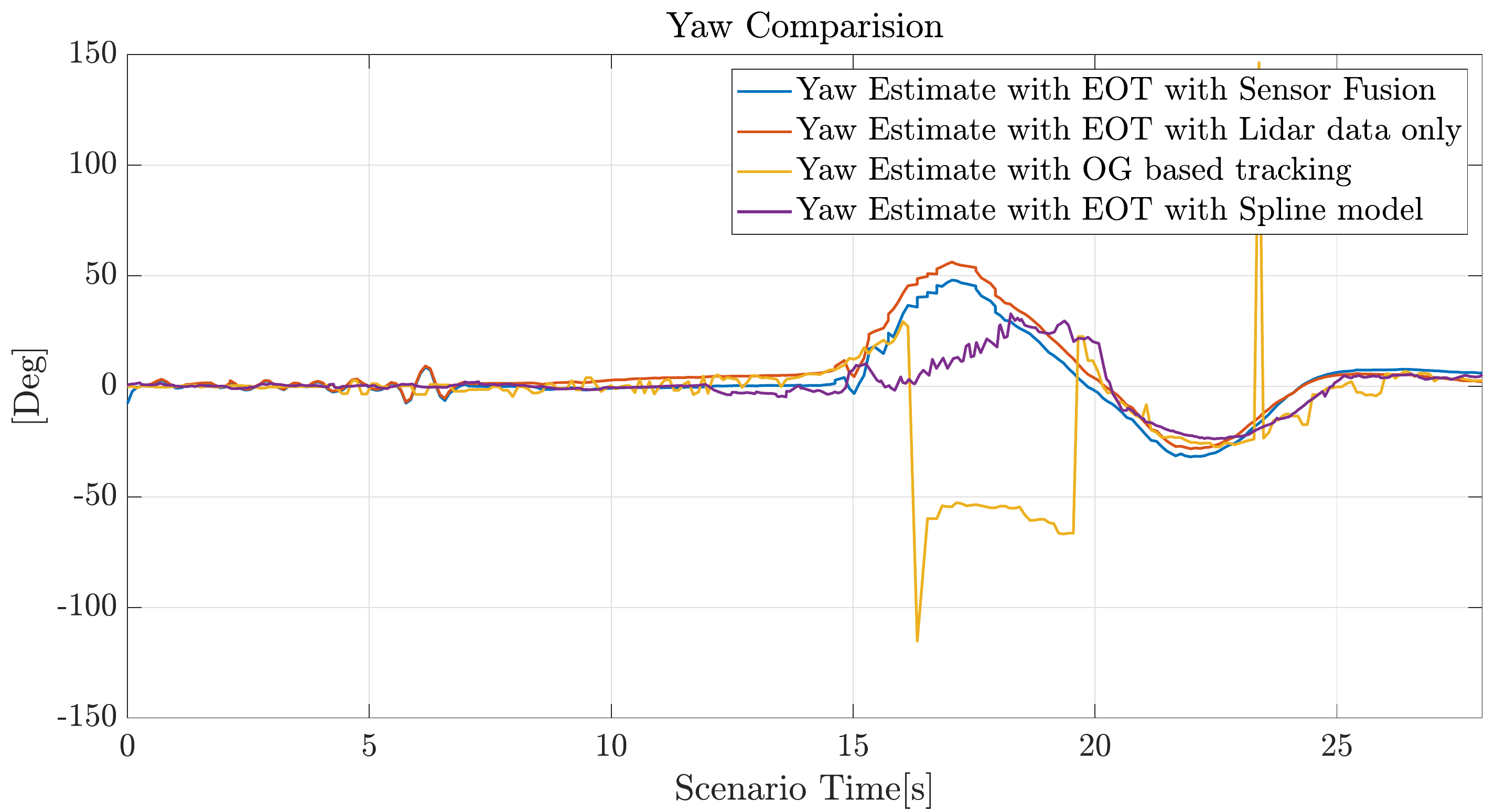}%
\label{fig:yaw_comp}}
\hfil
\subfloat[Comparison of shape parameters estimates ]{\includegraphics[width=0.5\linewidth]{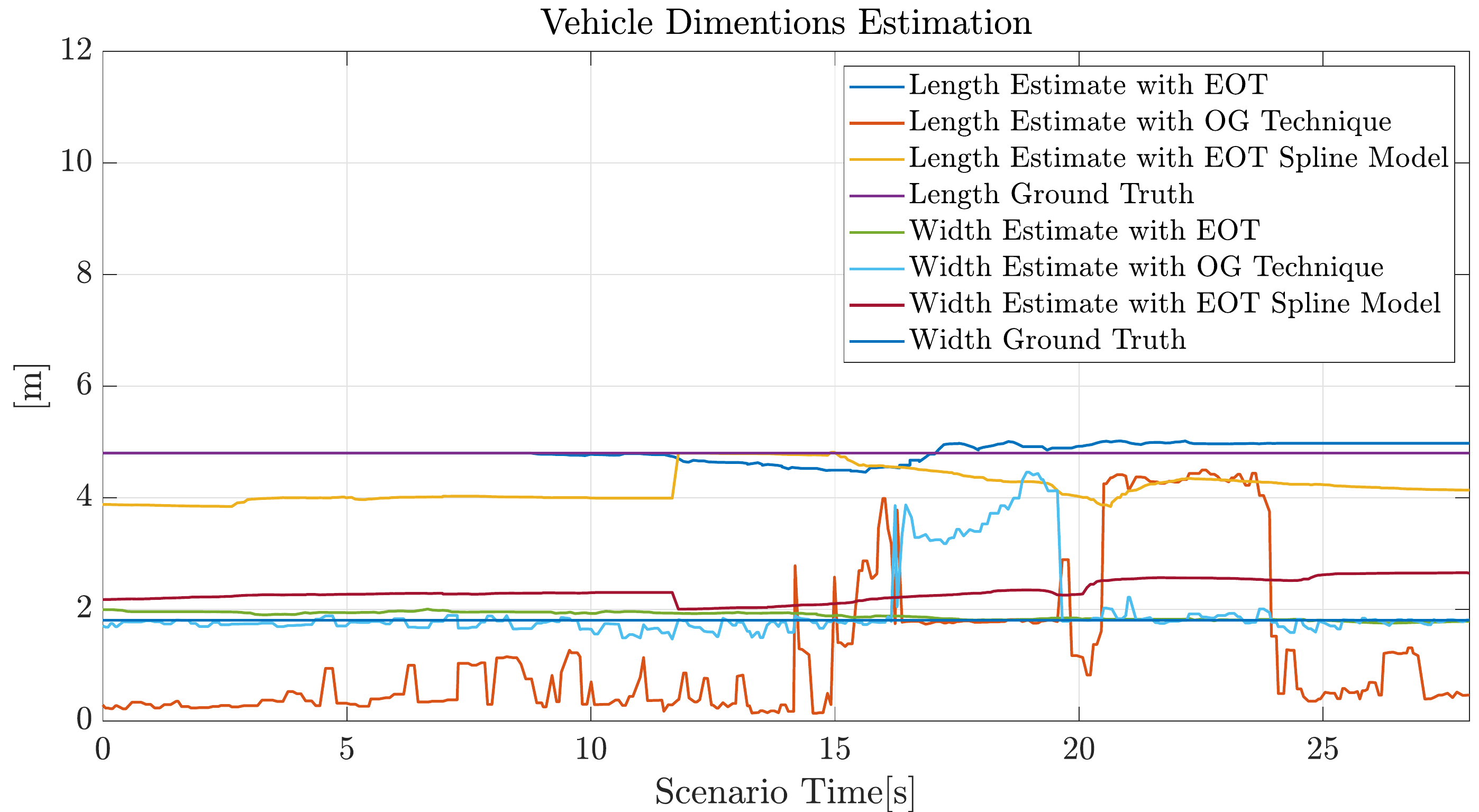}%
\label{fig:shape_comp}}
\caption{Yaw angle and shape parameters estimates obtained from different tracking approaches with ground truth for Scenario 1}
\label{fig:comparisions}
\end{figure*}

The scenario under investigation here is taken from the ``Variante Del Roggia'' segment of the track. We choose this scenario because of its significantly varying road curvature, as can be observed in Fig.~\ref{fig:Monza}. The results of the experimental scenario are made available in GIF format \href{https://polimi365-my.sharepoint.com/:i:/g/personal/10622973_polimi_it/Ebh06RjtzNdEhkkiXBKlt-UBoGjvPyPNiV0Qy45gGaAtzw?e=LQsUdK}{here} for further clarity. Additionally, results corresponding to the scenario at '' Variante Del Rettifilo '' 
are attached \href{https://polimi365-my.sharepoint.com/:i:/g/personal/10622973_polimi_it/EVW31PY3Dn9Dk0ginEnn4AUBvEsnoXUNNs_m0EXTj2v4dg?e=KC1yHu}{here} as well. 
In this scenario under analysis, an object-following maneuver is developed while the ego vehicle sees single and double sides of the object, necessitating the switch between single side and double sides measurement models. The velocities of the vehicles are around $18 m/s$ during this experiment.

We performed a comparative study with various recent algorithms to validate our proposed approach. In our previous work, \cite{2021_Integrated}, a Point Object Tracking~(POT) implementation of Global Nearest Neighbour~(GNN) filter with UKF estimator, object states are computed in curvilinear coordinates. However, with POT assumption, no object's shape and yaw angle estimates are performed. The algorithm is able to provide object estimates at $20 Hz$. State estimates using Occupancy Grid~(OG) based tracking technique, \cite{mentasti2021algorithms} provides object shape as well as yaw angle estimate and operates at $10 Hz$. A GMPHD filtering recursion with UKF estimator using spline measurement model proposed by \cite{2018_Spline_Measurement_Model} is also developed for validation of the proposed approach. States estimates are obtained at a frequency of $6 Hz$. Furthermore, the results obtained with two different setups of the proposed algorithm, first using Lidar data only as input and second with fused data from both Lidar and Radar sensors are compared.

In Fig.~\ref{fig:position_comp}, the results of this scenario are illustrated. These comparisons are made in VRF ($\textbf{E}$) based on cartesian coordinates because of the unavailability of ground truth in curvilinear coordinates. Estimated object states obtained from our proposed algorithm and \cite{2021_Integrated} are converted to cartesian coordinates for comparing with the ground truth. The ground truth values in the cartesian coordinates are obtained from an RTK-GPS  sensor installed in the tracked vehicle. The proposed algorithm outperforms the other algorithms in terms of positional accuracy of the estimated state. A root mean square error of $0.401 m$ and $0.51 m$ were observed respectively for the setup with and without sensor fusion for the proposed algorithm.

Fig.~\ref{fig:yaw_comp} illustrates the results obtained from different algorithms for yaw angle estimates. Due to the unavailability of the ground truth values for the yaw angle, results from different algorithms are compared among themselves. Visual assessment of the scene showed that the proposed algorithm can estimate the yaw angle with reasonable accuracy. The fluctuation of the yaw estimate from the OG-based technique is due to the nature of the detection obtained from the OG-based detector. As illustrated here, the algorithm cannot differentiate between the leading side of the bounding box and can provide wrong estimates. For the Spline measurement model-based EOT tracker, yaw angle estimates were observed to be erroneous due to the rapidly changing curvature of the road and only a single side of the tracked object being observed during this maneuver. Due to the integration of the road model and object state representation in curvilinear coordinates, estimation from the proposed algorithm is significantly better than the presented alternatives. 
\begin{figure}[t]
    \centering
    \includegraphics[width = 0.5\textwidth]{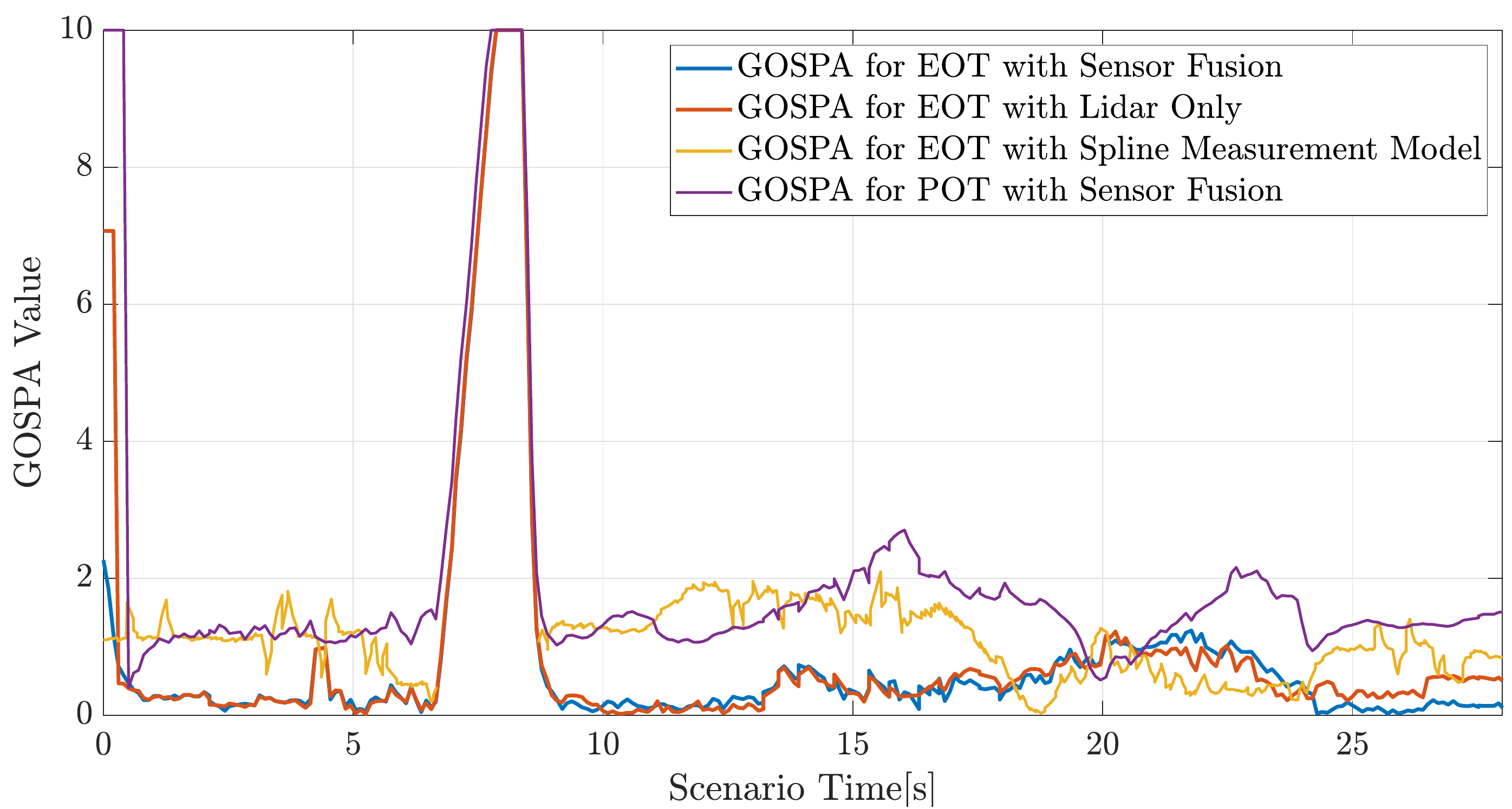}
    \caption{Comparision of GOSPA metric for different algorithms}
    \label{fig:GOSPA_comp}
\end{figure}
Fig.~\ref{fig:shape_comp} validates the extent estimates obtained from the proposed algorithm. The width estimated from the proposed EOT algorithm is confirmed with the outcome of the OG-based approach and the ground truth (approximately $2m$). Spline measurement model-based EOT tracker over estimates width at multiple instances. The length estimates from the OG-based approach, which can compute the length observed by the Lidar sensor only, strongly underestimates the length at multiple time instances. The proposed EOT algorithm's length estimates closely match the ground truth (approximately $4.8m$) and outperform the spline measurement model-based EOT algorithm estimates. \par
The proposed approach is presented as a filtering algorithm and hence is validated using the GOSPA metric. The GOSPA values obtained for the various algorithms are illustrated in the Fig.~\ref{fig:GOSPA_comp}. We observe that the use of sensor fusion for EOT provides better state estimates, and state representation in curvilinear coordinates reduces the localization error of the tracked object. The algorithm can also provide state estimates with reasonable GOSPA values except for the instances in between $7s$ to $9s$. The GOSPA performs poorly due to sudden drift in ground truth values obtained from the track GPS sensor. This drift can also be observed in the Fig.~\ref{fig:x_comp}.

\section{Conclusions and Future works} \label{Conclusion}
The key contribution of this work is to provide state estimation of dynamic objects within road bounds in curvilinear road coordinates. Extended object representation of these objects enables estimation of the kinematic and extent state. A GM-PHD Filter for EOT with a UKF estimator is used for this estimation process. A hybrid sensor fusion architecture consisting of Lidar and Radar is employed to obtain information regarding these objects. The algorithm is validated through simulation, and experimental data collected from the test runs at the Monza Eni. Circuit. A comparative study with the Occupancy-Grid-based tracking technique, Point Object Tracking in curvilinear coordinates, and EOT with spline-based measurement model is explored to validate the object state estimates obtained using the EOT algorithm. We also demonstrate that the state representation in curvilinear coordinates increases the accuracy in yaw angle estimate in high road curvature scenarios.\par
Future work will focus on generalizing the proposed approach, removing constraints on the object's shape. Tracking objects other than vehicles requires a different approach than predefined rectangular shapes to compute measurement generating points. Gaussian Processes, Splines models, etc., can be explored to develop star convex-shaped object models that provide free-form object shape.

\bibliographystyle{IEEEtran}
\bibliography{main.bbl}

\end{document}